%% file: arxiv.tex
\documentclass[10pt,twocolumn,letterpaper]{article}

\usepackage[pagenumbers]{iccv} 

\usepackage[accsupp]{axessibility} 
\usepackage{multirow}
\usepackage{amssymb}
\usepackage{pifont}
\usepackage{colortbl}
\usepackage{amsmath}
\usepackage{soul}
\usepackage{amsthm}
\usepackage[most]{tcolorbox}
\newtheorem{proposition}{Proposition}

\definecolor{iccvblue}{rgb}{0.21,0.49,0.74}
\usepackage[pagebackref,breaklinks,colorlinks,allcolors=iccvblue]{hyperref}

\def\paperID{6754} 
\def\confName{ICCV}
\def\confYear{2025}

\title{KinMo: Kinematic-aware Human Motion Understanding and Generation}

\author{Pengfei Zhang$^1$\footnotemark[1]\quad ~Pinxin Liu$^2$\footnotemark[1]\quad ~Pablo Garrido$^4$\quad ~Hyeongwoo Kim$^3$\quad ~Bindita Chaudhuri$^4$\footnotemark[2]\\
$^1$ \small University of California, Irvine, $^2$ University of Rochester, $^3$ Imperial College, London, $^4$ Flawless AI\\
$^1${\tt\small pengfz5@uci.edu}, $^2${\tt\small pliu23@u.rochester.edu}, \\
$^3${\tt\small hyeongwoo.kim@imperial.ac.uk}, $^4${\tt\small \{pablo.garrido, bindita.chaudhuri\}@flawlessai.com}
}

\begin{document}

\twocolumn[{%
\renewcommand\twocolumn[1][]{#1}%
\maketitle
\begin{center}
    \centering
    \vspace{-0.7cm}
    \captionsetup{type=figure}
    \includegraphics[width=\textwidth]{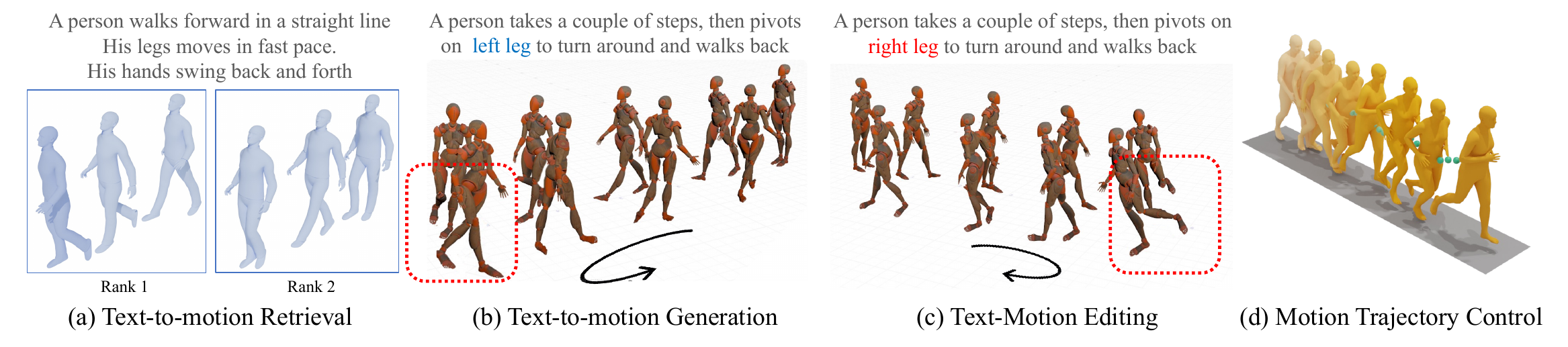}
    \vspace{-0.9cm}
    \captionof{figure}{\textbf{We present KinMo}, a method that achieves fine-grained motion understanding for (a) effective text-motion retrieval, and text-aligned motion (b) generation, (c) editing, and (d) trajectory control on local kinematic body parts.}
\label{Teaser}
\end{center}%
}]
\footnotetext[1]{* Work done as interns at Flawless AI. These authors contributed equally to this work. \dag ~Corresponding Author.}
\vspace{-0.3cm}
\input{sec/0_abstract}

\input{sec/1_intro}

\input{sec/2_related_works}
\input{arxiv_sec/3_method}

\input{arxiv_sec/4_experiment}
\input{sec/5_conclusion}

{
    \small
    \bibliographystyle{ieeenat_fullname}
    \bibliography{main}
}

\newpage

\input{arxiv_sec/supp}

\end{document}

%% file: sec/0_abstract.tex
\begin{abstract}
%
Current human motion synthesis frameworks rely on global action descriptions, creating a modality gap that limits both motion understanding and generation capabilities. 
A single coarse description, such as ``run", fails to capture details such as variations in speed, limb positioning, and kinematic dynamics, leading to ambiguities between text and motion modalities. 
To address this challenge, we introduce \textbf{KinMo}, a unified framework built on a hierarchical describable motion representation that extends beyond global actions by incorporating kinematic group movements and their interactions.
We design an automated annotation pipeline to generate high-quality, fine-grained descriptions for this decomposition, resulting in the KinMo dataset and offering a scalable and cost-efficient solution for dataset enrichment.
To leverage these structured descriptions, we propose Hierarchical Text-Motion Alignment that progressively integrates additional motion details, thereby improving semantic motion understanding.
Furthermore, we introduce a coarse-to-fine motion generation procedure to leverage enhanced spatial understanding to improve motion synthesis. 
Experimental results show that KinMo significantly improves motion understanding, demonstrated by enhanced text-motion retrieval performance and enabling more fine-grained motion generation and editing capabilities.
Project Page: {\small\url{https://andypinxinliu.github.io/KinMo}}
\end{abstract}

%% file: sec/1_intro.tex
\section{Introduction}
\label{sec:intro}

Controlling human motion through natural language is a rapidly expanding area within computer vision, enabling interactive systems to generate or modify 3D human motions based on textual input. This technology has a wide range of applications, including robotics\cite{koppula2013anticipating}, digital avatar\cite{li2024isolated, huang2024modelingdrivinghumanbody, adaptive}, and automatic animation \cite{zhu2024champ,zhang2024handformer2t,tang2025generative, gaussianstyle, song2024tri}, where human-like motion is crucial for user interaction and immersion.
Despite efforts to generate general motion~\cite{shafir2023human,tevet2022humanmotiondiffusionmodelmdm,zhang2022motiondiffuse,sun2024diffposetalk,pinyoanuntapong2024mmm,guo2024momask}, fine-grained control over individual body parts remains largely an unsolved challenge. Current models are proficient in producing coherent whole-body movements from global action descriptions but struggle when tasked with controlling local body parts independently. This limitation prevents them from achieving precision and adaptability for real-world applications.

Recent advances~\cite{petrovich2024multi,huang2024como} have introduced more refined approaches by incorporating controllability into motion generation. However, these models are limited to processing simple instructions and lack the compatibility required for scenarios where multiple body parts must coordinate to perform complex actions. Similarly, generative models for motion synthesis~\cite{guo2024momask, pinyoanuntapong2024mmm} present innovative methods but do not directly address the issue of controlling specific body parts with specific textual descriptions.

This challenge stems from the inherent ambiguity of motion text descriptions in existing datasets.
For example, multiple phrases (such as \textit{pick up an object from the ground} and \textit{bend down to reach something}) can describe the same motion. In contrast, a single term (such as \textit{running}) can encompass a wide range of variations, depending on factors such as speed, arm movement, or direction. This many-to-many mapping problem~\cite{liu2023bridging} hinders existing models from handling the multiplicity of natural language or motions, often resulting in inconsistent or unnatural outcomes when trying to generate or edit specific body parts.


To solve this problem, we introduce a novel motion representation based on six fundamental kinematic components: torso, head, left arm, right arm, left leg, and right leg. 
Unlike existing methods that treat the body as a whole, our approach explicitly models each component and its interactions, enabling a more detailed representation of global action through localized body movements.
For instance, a \textit{sneaking} motion should involve coordinated torso and leg movement, while the arms are used for balance.
Based on this, we propose a kinematic-aware formulation that opens new possibilities for text-motion understanding, fine-grained motion generation, and editing.
Building on this insight, we propose a kinematic-aware formulation, which enables improved text-motion understanding, fine-grained motion generation, and editing capabilities.

To achieve this, we reformulate existing motion representations and enhance the widely used HumanML3D~\cite{guo2022humanml3d} dataset by introducing a semi-supervised annotation system that enriches motion data with body-part-specific descriptions, forming our KinMo dataset.
We investigate the retrieval capability of these body-part-specific descriptions, demonstrating that our proposed Hierarchical Text-Motion Alignment effectively integrates these semantics to further enhance text-motion understanding.
In addition, we demonstrate how this enhanced understanding benefits motion generation by extending the MoMask~\cite{guo2024momask} model to support fine-grained body-part generation and editing, enabling greater control and manipulation of motion sequences.
Our contributions can be summarized as follows:

\begin{enumerate} \setlength{\parskip}{0em} 
    \item We introduce \textbf{KinMo}, a novel framework that decomposes human motion through a three-level hierarchy: global actions, local kinematic groups, and group interactions. This hierarchical representation significantly bridges the gap between text and motion. We further develop a semi-supervised annotation pipeline for generating our dataset.
    
    \item We propose a Hierarchical Text-Motion Alignment method that leverages enriched textual descriptions by progressively encoding them and integrating hierarchical semantics. This approach enhances retrieval capabilities, showing notable improvements in semantic motion understanding within spatial contexts.

    \item We extend \textbf{Motion Generation} process into a coarse-to-fine procedure, which enhances motion understanding by transitioning from global actions to joint groups and their interactions. This supports various generative and editing applications with fine-grained control.

\end{enumerate}

%% file: sec/2_related_works.tex
\section{Related Work}

\noindent\textbf{Text-to-Motion Understanding.} 
Similar to other modality alignments~\cite{oord2019representationlearningcontrastivepredictive,yang2024dgl,liu2025contextualgesturecospeechgesture,liu2025intentionalgesturedeliverintentions}, alignment/retrieval between text and motion modalities is the key indicator of motion understanding. PoseScript~\cite{delmas2022posescript} uses fine-grained text descriptions to represent various human poses. MotionCLIP~\cite{tevet2022motionclip} and TMR~\cite{petrovich2023tmr} enhance the alignment from single poses to motion sequences. MotionLLM~\cite{chen2024motionllm} creates a large corpus of Motion-QA for text-motion understandings. 
However, these methods only focus on global action descriptions, ignoring the extent of local kinematic movements, which are essential for alignment and motion understanding.

\noindent\textbf{Text-to-Motion Generation.} 
Diffusion Models have been the main trend for various modalities~\cite{Lu_2023_ICCV,lu2024mace,lu2023tf,ning2024dctdiff,huang2024scalingconcepttextguideddiffusion,huang2024highqualityvisuallyguidedsoundseparation,lu2024robust,gao2024eraseanything,li2025setstraightautosteeringdenoising} and demonstrated notable success in motion generation~\cite{tevet2022humanmotiondiffusionmodelmdm, zhang2022motiondiffuse, liu2025gesturelsmlatentshortcutbased}. T2M-GPT~\cite{zhang2023generating} and MotionGPT~\cite{ribeiro2024motiongpt} represent motions as discrete tokens and leverage autoregressive models to improve motion generation quality. Masking Motion Models~\cite{pinyoanuntapong2024mmm, guo2024momask, liu2025contextualgesturecospeechgesture} further improve motion generation quality with a bidirectional masking mechanism.

Large Language Models (LLMs) have empowered various understanding and generation with fine-grained control~\cite{tang2024vidcomposition,tang2024empowering, liang2024languageguidedjointaudiovisualediting}. LGTM \cite{lgtm} and FG-MDM~\cite{fgmdm} use LLMs to generate additional texts to describe local motions to assist the generation process. 
Although these methods partially focus on fine-grained local motion control, they cannot address the core ambiguity problem between the two modalities. To address this problem, we take a step further in reformulating a linguistically describable motion representation from global actions to groups and interactions.

\noindent\textbf{Trajectory Control and Editing.}
Diffusion-based models~\cite{tevet2022humanmotiondiffusionmodelmdm, zhang2022motiondiffuse, athanasiou2024motionfix} can perform zero-shot editing by infilling specific joints. TLControl~\cite{tlcontrol}, OmniControl~\cite{omnicontrol}, and CoMo~\cite{huang2024como} can control arbitrary joints at any time by combining spatial and temporal control together. 
However, none of these methods adopt a fine-grained approach that allows local editing while ensuring overall motion compatibility.

%% file: arxiv_sec/3_method.tex
\begin{figure*}
\centering
\includegraphics[width=0.95\linewidth]{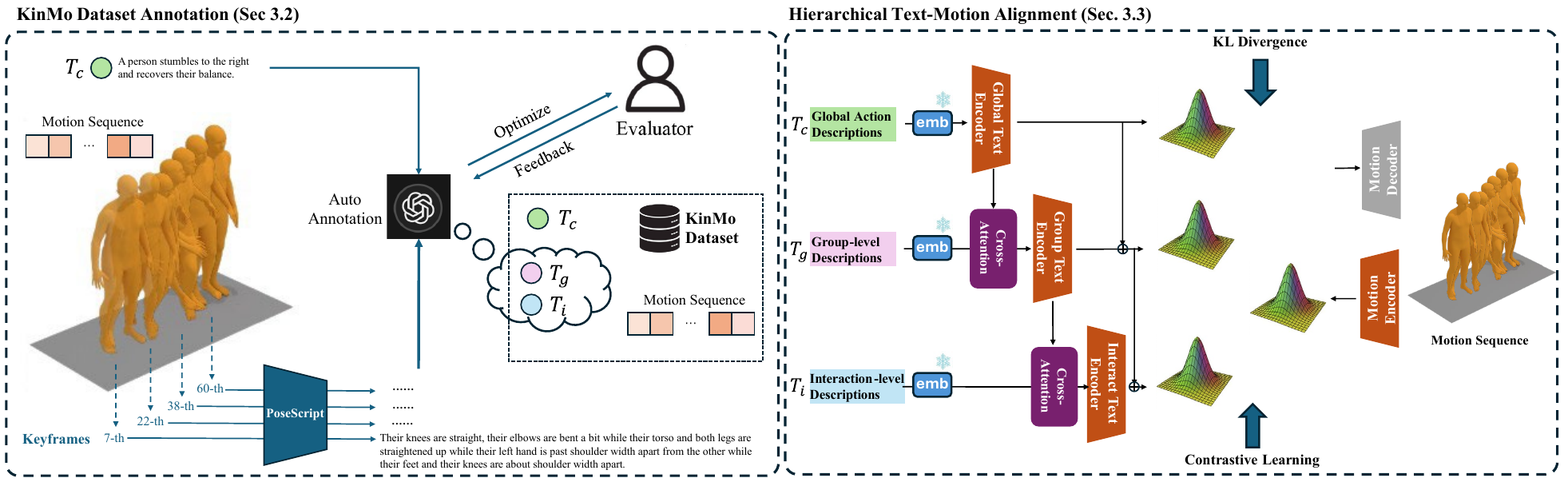}
\vspace{-1em}
\caption{\textbf{KinMo Framework}. \textit{Left:} We extract pose descriptions of the keyframes and feed them into an LLM to produce group- and interaction-level descriptions, which generate KinMo dataset together with original motion sequences and global action texts.
\textit{Right}: We apply encoders with the same architecture (brown) to process the features of global action, group-level descriptions, and interaction-level descriptions extracted from a pretrained model (blue: emb). 
The cross-attention layer (purple) is employed to combine embeddings of different levels to enable hierarchical representation learning, with contrastive learning at each level for modality alignment.
}
\vspace{-1.5em}
\label{fig:framework}
\end{figure*}


\section{Kinematic-aware Human Motion}
In this section, we first introduce the core principle of KinMo, which bridges the gap between text and motion by linearly transforming motions into \textbf{linguistically describable representations} in 3D space (Sec.~\ref{sec:representation}). We then propose an LLM-based pipeline to annotate these representations and present the Kinematic-aware Motion-Text (KinMo) dataset (Sec.~\ref{sec:dataset}). Additionally, we propose an alignment method to achieve spatial understanding given the enriched textual descriptions (Sec.~\ref{sec:understanding}).

\subsection{Describable Motion Representations}
\label{sec:representation}

\noindent\textbf{Existing Motion Representations.} Current text-motion alignment research \cite{guo2022humanml3d,petrovich2023tmr,guo2024momask,pinyoanuntapong2024mmm,petrovich2022temos,petrovich24stmc} represents motion as the time evolution of each body joint \(j\in J\) of the human body, characterized by its position \(\mathbf{p}_j(t)\), axis-angle rotation relative to its parent in the kinematic tree \(\mathbf{r}_j(t)\), and relative angular velocity \( \mathbf{v}_j(t) \) with respect to the center joint.\footnote{W.l.o.g. we omit some boundary conditions, e.g., foot contact and center joint selection, as they do not affect the core conclusions. The supplementary document (Appendix \ref{conversion_proof}) shows a detailed computation of existing motion representations.} 
However, this representation is hard to describe in natural language.
Besides, global action descriptions struggle to represent local movements.
To solve this problem, we create an intermediate representation of a motion that is explicitly describable in natural language. 
Specifically, we reformulate motion representations by organizing joints into a set of \textbf{kinematic groups} following kinematic tree, defined as \(G=\{\text{Torso, Neck, Left Arm, Right Arm, Left Leg, Right Leg}\}\), where each group \( g\in G \) consists of joints \( J_g \subseteq J \).
%

\vspace{0.1cm}
\noindent\textbf{Kinematic-Group Representations.}
For each group \( g \) at time \( t \), we define the Group Position \( \mathbf{P}_g(t) \) as the average position \(\mathbf{p}_j(t)\) of the joints within that group:
\vspace{-0.3cm}
\begin{equation}
    \mathbf{P}_g(t) = \frac{1}{|J_g|} \sum_{j \in J_g} \mathbf{p}_j(t)\text{.}
\end{equation}
\vspace{-0.1cm}
\noindent We then define the \textit{Limb Angles} \( \Theta_g(t) \) as the collection of joint rotations \(\mathbf{r}_j(t)\) within the group, and the \textit{Group Velocity} \( \mathbf{V}_g(t) \) as the average 
velocity \(\mathbf{v}_j(t)\) of the joints:
\begin{equation}
    \mathbf{\Theta}_g(t) = \{ \mathbf{r}_j(t) \mid j \in J_g \}, \quad
    \mathbf{V}_g(t) = \frac{1}{|J_g|} \sum_{j \in J_g} \mathbf{v}_j(t)\text{.}
\end{equation}
\noindent\textbf{Group-Interaction Representations.} Human motion also involves the relationships between each pair of groups \( (g, h) \in G \times G \). 
\begin{equation}
    \begin{bmatrix}
        \Delta \mathbf{P}_{g,h}(t) \\
        \Delta \mathbf{\Theta}_{g,h}(t) \\
        \Delta \mathbf{V}_{g,h}(t)
    \end{bmatrix}
    =
    \begin{bmatrix}
        \mathbf{P}_h(t) - \mathbf{P}_g(t) \\
        \mathbf{\Theta}_{h \cap g}(t) \\
        \mathbf{V}_h(t) - \mathbf{V}_g(t); \mathbf{v}_{h \cap g}(t)
    \end{bmatrix}\text{,}
\end{equation}
where \( \Delta \mathbf{P}_{g,h}(t) \) denotes the difference in position, \( \Delta \mathbf{\Theta}_{g,h}(t) \) represents the angles at the connecting joint (if exists), 
and \( \Delta \mathbf{V}_{g,h}(t) \) is the relative angular velocity and the angular velocity at the connecting joint (if exists) between two groups. 

The proposed formulation of motion representations is a linear transformation of the existing formulation used in current text-motion alignment methods \cite{petrovich2023tmr,guo2024momask,pinyoanuntapong2024mmm,petrovich2022temos,petrovich24stmc} and can be transformed back to existing representations, as detailed in Appendix \ref{conversion_proof}. Additionally, our formulation is inherently compatible with natural language descriptions, capturing both the movements of individual kinematic groups and their interactions. With this formulation, the key task is to annotate our proposed linguistically describable motion representations to collect textual descriptions of kinematic groups and group interactions.


\subsection{KinMo Dataset}
\label{sec:dataset}

\noindent\textbf{Kinematic-aware Joint-Motion Text Annotation.}
Good annotations of the proposed motion representations must capture both spatial and temporal details of each kinematic group and their interactions. We employ a two-step LLM-based strategy to ensure high-quality automatic annotation, as shown in \cref{fig:framework}.
Using this strategy, we enhance the HumanML3D dataset~\cite{guo2022humanml3d} with fine-grained annotations.

\vspace{0.1cm}
\noindent\textbf{Spatial-Temporal Motion Processing.}
A motion consists of a sequence of pose frames over time. Existing human motion understanding models~\cite{chen2024motionllm, guo2022humanml3d, ribeiro2024motiongpt} struggle to capture subtle local movements due to the complex interplay of spatial and temporal dynamics. To address this, we propose a two-stage disentanglement approach, first resolving spatial dynamics and then temporal dynamics.

To extract detailed spatial information, we adopt PoseScript~\cite{delmas2022posescript} to generate annotations for each pose frame, capturing precise angular rotations of joints for any given human pose. 
To obtain fine-grained temporal information, we propose a keyframe selection pipeline. We use sBERT~\cite{reimers2019sentence} to extract embeddings for the per-frame pose descriptions. We assess the similarity of poses across the time frames by calculating the cosine similarity between text embeddings.
If the cosine similarity falls below a user-defined threshold, we label that frame as a keyframe.
The pose differences within each kinematic group over a specified time window are used to approximate local temporal motions during that period.

\vspace{0.1cm}
\noindent\textbf{Semi-supervised Annotation.}
We then design an automatic annotator using GPT-4o-mini~\cite{openai2024gpt4technicalreport} to generate textual descriptions of kinematic groups and their interactions based on keyframe pose annotations. 
To refine the prompt, two human evaluators iteratively assess and improve the annotation process. We begin by randomly sampling 20 preprocessed motion sequences and using GPT-4o-mini to infer textual descriptions based on keyframe pose annotations. The two evaluators independently review the generated descriptions, documenting errors made by the LLM. These insights are then fed back into the model to refine the prompt design. This iterative process continues until the evaluators reach a consensus, achieving a kappa statistic above 0.8. At this point, we ensure that subsequent LLM-generated descriptions for the remaining dataset align with our objectives. Additional details on the text prompt and example descriptions can be found in Appendix \ref{sec:sub_dataset}.

In summary, the KinMo dataset provides three types of descriptions for each motion: 1) Global action descriptions; 2) Spatial and temporal per-group descriptions of the movement and dynamics of each kinematic group \(g\); 3) Spatial and temporal per-group-pair descriptions of the relative movements and dynamics between each pair of kinematic groups \(h\) and \(g\).
We will refer to them as \textit{global action descriptions} (\(T_c\)), \textit{group-level descriptions} (\(T_g\)), and \textit{interaction-level descriptions} (\(T_i\)), respectively.

\subsection{Hierarchical Text-Motion Alignment (HTMA)}
\label{sec:understanding}

Existing text-motion alignment methods typically encode global actions and parse additional descriptions directly~\cite{fgmdm,jin2024localactionguidedmotiondiffusion,finemogen}, which limits spatial understanding. Rich contextual dependencies and explicit positional relationships among descriptions make it challenging to capture high-quality semantics when directly encoding additional descriptive levels~\cite{parco,liu2023bridging}. Instead, we propose a hierarchical framework that first encodes the global action as \textbf{coarse embeddings} and then progressively incorporates group- and interaction-level descriptions to refine the embeddings, achieving hierarchical alignment, as illustrated in Figure~\ref{fig:framework}.

\vspace{0.1cm}
\noindent\textbf{Modality Encoders.} To align motion and text modalities, we follow TMR~\cite{petrovich2023tmr} to map textual descriptions and motion into a shared co-embedding space. Motion and text encoders are Transformer-based~\cite{transformer} with additional learnable distribution parameters, as in the VAE-based ACTOR model~\cite{petrovich2021actionconditioned3dhumanmotion}. We follow a probabilistic approach that utilizes two prefix tokens for each text or motion sequence to learn the \((\mu, \Sigma)\) of a Gaussian distribution \(\mathcal{N}\), from which a latent vector \(z \in \mathbb{R}^d\) is sampled.

Motion sequences are processed directly by the motion encoder.
For the textual inputs, we first extract text features from a pre-trained and frozen RoBERTa~\cite{roberta} or DistilBERT~\cite{sanh2019distilbert} to obtain global action, group-level, and interaction-level text descriptions, and then encode these features with cross-attention in a hierarchical process:

\vspace{-5pt}
\begin{equation}
\mathbf{h_c} = E_c\big(\text{emb}(T_{c})\big)\text{,}
\end{equation}
\vspace{-5pt}
\begin{equation} \label{eq:group}
\mathbf{h_g} = E_g\big(\text{CrossAttn}(\text{emb}(T_{g}), \mathbf{h_c})\big)\text{,}
\end{equation}
\vspace{-5pt}
\begin{equation}
\mathbf{h_i} = E_i\big(\text{CrossAttn}(\text{emb}(T_{i}), \mathbf{h_g})\big)\text{,}
\end{equation}
where $\text{emb}(\cdot)$ represents a pre-trained RoBERTa or DistilBERT model for feature extraction, and $E_{{c, g, i}}$ are the VAE-based ACTOR models used as text encoders for each level. CrossAttn$(\cdot, \cdot)$ denotes a single cross-attention layer with a residual connection. The cross-attention mechanism allows each level to build upon previous embeddings, creating progressively refined semantic embeddings. The group-level embeddings incorporate global action, while the interaction-level embeddings incorporate the group-level.\looseness=-1


%
\begin{figure*}
\centering
\includegraphics[width=\linewidth]{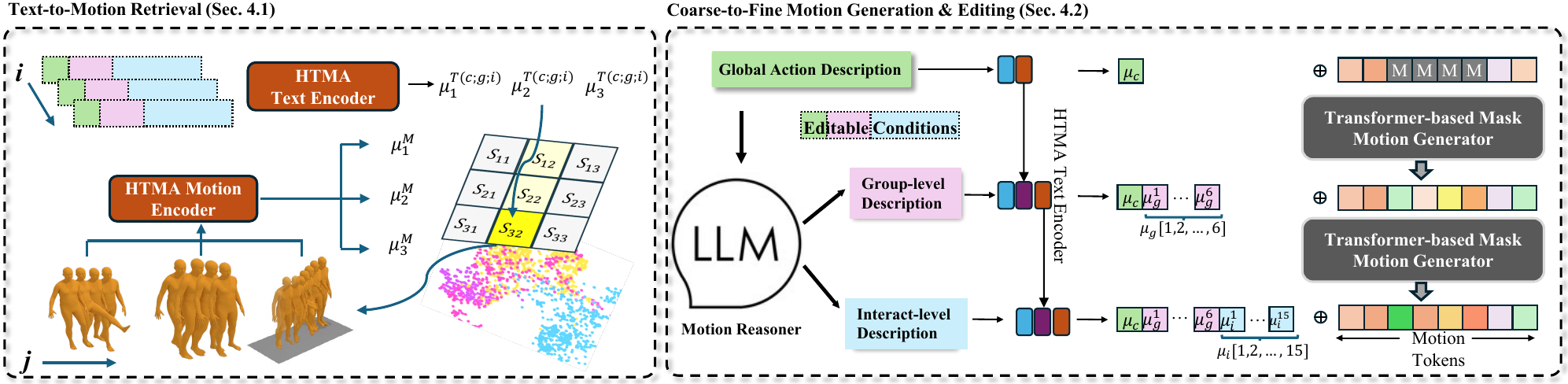}
\vspace{-2em}
\caption{ \textbf{Motion Retrieval and Generation.} \textit{Left}: Overview of Text-to-Motion Retrieval, where we compute the similarity matrix defined between text and motion embeddings. Here, we present a batch of three samples with an example. To retrieve the most similar motion to the 2nd text \(T_{c_2}\),  
we use \textit{Motion Reasoner} to generate corresponding group- and interaction-level descriptions, producing aligned embeddings \(\mu^{T(c;g;i)}_2\). We then check the similarity matrix against the motion embeddings, where \(S_{32}\) has the highest value, indicating that the 3rd motion is the closest match to the 2nd text. \textit{Right:} Given a global action, \textit{Motion Reasoner} generates group- and interaction-level descriptions. The aligned text embeddings at each level are prepended to the motion tokens and collectively fed into the motion generator through a coarse-to-fine generation process. Editing can be performed on the texts either before or after the Motion Reasoner stage. \textit{HTMA Text/Motion Encoder (\cref{sec:understanding})} aligns text and motion into a shared embedding space to obtain their respective aligned representations.}

\vspace{-1.5em}
\label{fig:motiongeneration}
\end{figure*}
%

\vspace{0.1cm}
\noindent\textbf{Contrastive Learning.} We use contrastive learning to align the embeddings of motion and text modalities in each hierarchy \cite{petrovich2023tmr}. For simplicity, we denote any level of text and motion pair as \((z^T, z^M),T\in\{T_c,T_g,T_i\}\)
. For a batch of \( N \) positive pairs \((z_1^T, z_1^M), \ldots, (z_N^T, z_N^M)\), any pair \((z_i^T, z_j^M)\) where \( i \neq j \) is considered a negative sample. The similarity matrix \( S \) computes the pairwise cosine similarities for all pairs in the batch, defined as \( S_{ij} = \text{cos}(z_i^T, z_j^M) \). We apply an InfoNCE loss~\cite{oord2019representationlearningcontrastivepredictive}, as follows:
{\small
\begin{equation}
\begin{split} 
    \mathcal{L}_{\text{NCE}} &= \frac{-1}{2N} \sum_{T} \sum_{i} \left( \log \frac{\exp{S_{ii}/\tau}}{\sum_{j} \exp{S_{ij}/\tau}} + \log \frac{\exp{S_{ii}/\tau}}{\sum_{j} \exp{S_{ji}/\tau}} \right)\text{,}
\end{split}
\label{eq:infonce}
\end{equation}
}
where \( \tau \) represents a temperature parameter.

To maximize the proximity between the two modalities, we follow TMR~\cite{petrovich2023tmr} to construct a weighted sum of 3 losses: (a) Kullback–Leibler divergence loss \(\mathcal{L}_{\text{KL}}\), (b) cross-modal embedding similarity loss \(\mathcal{L}_{\text{E}}\), and (c) motion reconstruction loss \(\mathcal{L}_{\text{R}}\) for each semantic hierarchy.

\section{Motion Understanding and Generation}
\label{motion_understanding_generation}
KinMo framework provides new insight into motions. By bridging the gap between text and motion through our proposed motion representations and alignment method, we will show how KinMo achieves motion understanding\footnote{Since \textit{motion understanding} currently lacks diverse downstream tasks, with only text-motion retrieval widely used, our claim of \textit{motion understanding} may be an overstatement. Here, we specifically explore its impact on text-motion retrieval.} and generation, as well as downstream applications (Sec.~\ref{sec:exp-applications}).


\noindent\textbf{Motion Reasoner.} Our goal is to generate group- and interaction-level descriptions based on global action inputs. We finetune LLaMA-3~\cite{grattafiori2024llama3herdmodels} on the KinMo Dataset to serve as a motion reasoner. The model is trained using the standard next-token prediction loss, with an added conditioning on global action $T_{c}$ to output the corresponding group-level descriptions $T_g$ and interaction-level descriptions $T_i$:
\begin{equation}
\mathcal{L}_{reasoner} = -\sum_{i=1}^{N} y_i \log(\hat{y}_i | T_{c}, T_{<i}),
\end{equation}
where $y_i$ and $\hat{y}_i$ represent the ground truth and predicted tokens at position $i$, respectively. $T_{<i}$ denotes the previously generated tokens. This loss encourages the model to generate motion descriptions at different granularity levels. 

\noindent\textbf{Text Alignment.} Given the additional descriptions of Motion Reasoner, we obtain their corresponding text embeddings \(\mu_c,\mu_g,\mu_i\) based on hierarchical text encoders from Sec.~\ref{sec:understanding} for various applications.

\subsection{Text-Motion Retrieval}

As shown in \cref{fig:motiongeneration}, for a given motion \(M\) and its corresponding text descriptions \(T_c,T_g,T_i\), the mean token of the output motion parameters \(\mu^M_j\) serves as aligned motion embedding, while the mean token of the output text parameters \(\mu^{T(c;g;i)}_i=[\mu_c;\mu_g;\mu_i]\) acts as aligned text embedding in the co-embedding space at different hierarchical levels. For retrieval, we directly compare these embeddings, identifying the best match by maximizing the cosine similarity between motion and text embeddings.

\vspace{-0.2cm}
\subsection{Coarse-to-Fine Motion Generation}

\label{sec:generation}
For motion generation, we adopt MoMask~\cite{guo2024momask} as the base architecture. Specifically, a VQ-VAE~\cite{vqvae,zhang2023generating} is trained to convert a motion sequence into discrete tokens. Then, given a text \(T_c\) prepended to the discrete motion tokens as input condition, a Transformer-based generator is trained with a masking-based strategy for token generation. The generated token will be used to query VQ-VAE to generate motions. To incorporate KinMo into the existing motion generation framework, we leverage Motion Reasoner to output group-level \(T_g\) and interaction-level \(T_i\) descriptions given input global action descriptions and then encode them into \(\mu_c,\mu_g,\mu_i\). In addition, we propose a coarse-to-fine generation procedure conditioned on the text hierarchy.

\vspace{0.1cm}
\noindent\textbf{Hierarchical Generation.}
As shown in \cref{fig:motiongeneration}, we extend MoMask~\cite{guo2024momask} from generation under condition \(\text{CLIP}(T_{\text{c}})\), into condition \(\mu_c,\mu_g,\mu_i\), which are aligned embeddings of \([T_{\text{c}}; T_{\text{g}}; T_{\text{i}}]\), through a coarse-to-fine generation process. Specifically, after the initial motion tokens are generated conditioned on \(\mu_{\text{c}}\), they will be re-fed into the generator to output intermediate tokens conditioned on \(\mu_{\text{g}}\). Then, the intermediate tokens are re-fed into the generator to produce the final tokens conditioned on \(\mu_{\text{i}}\). The final tokens are used to query motions into a trained VQ-VAE. For efficiency, the generator shares weights with the same logit classification loss functions that were used to reconstruct motion tokens for the three levels of text conditioning. Other configurations are the same as in MoMask~\cite{guo2024momask}.



%% file: arxiv_sec/4_experiment.tex
\section{Experiments}
\label{sec:experiment}

\begin{figure*}[thb]
\centering
\includegraphics[width=.95\linewidth]{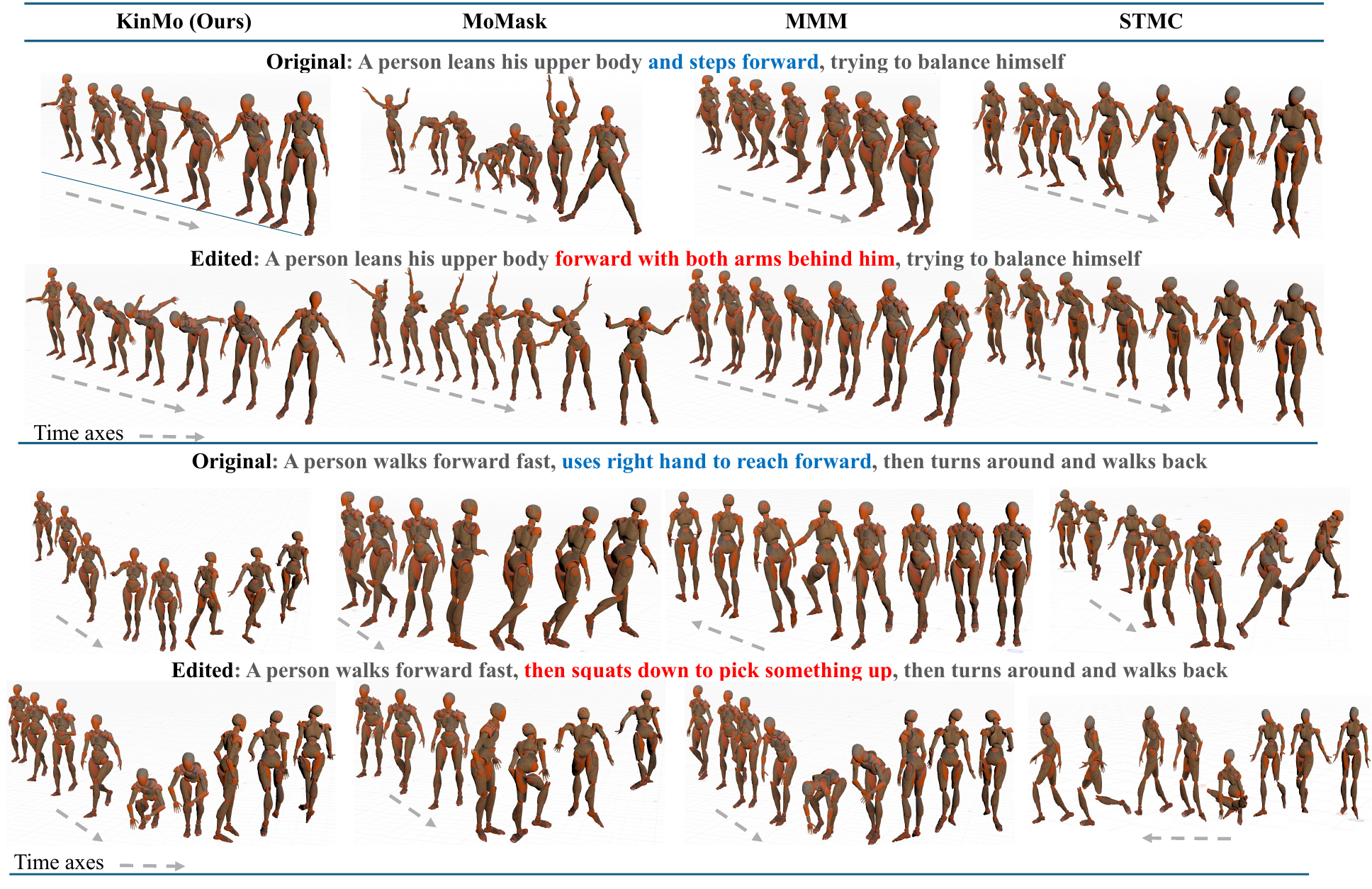}
\vspace{-0.53cm}
\caption{\textbf{Comparisons with different methods for two original and edited text descriptions from HumanML3D test set}. Unlike previous methods, our results match the input text descriptions better and show the ability to edit specific body parts.}
 \vspace{-0.2cm}
\label{fig:visual_comparison}
\end{figure*}

\begin{table*}[htb]
\centering
\caption{\textbf{Text-to-motion retrieval benchmark on HumanML3D.}
Evaluation protocols with decreasing difficulty from (a) to
(d).
}
\vspace{-10pt}
\label{tab:benchmark_h3d}
\resizebox{0.95\linewidth}{!}{
\begin{tabular}{ll|cccccc|cccccc}
\toprule
\multirow{2}{*}{\textbf{Protocol}} & \multirow{2}{*}{\textbf{Methods}} & \multicolumn{6}{c|}{Text-motion retrieval} & \multicolumn{6}{c}{Motion-text retrieval} \\
& & \small{R@1 $\uparrow$} & \small{R@2 $\uparrow$} & \small{R@3 $\uparrow$} &  \small{R@5 $\uparrow$} & \small{R@10 $\uparrow$} & \small{MedR $\downarrow$} & \small{R@1 $\uparrow$} & \small{R@2 $\uparrow$} & \small{R@3 $\uparrow$} & \small{R@5 $\uparrow$} & \small{R@10 $\uparrow$} & \small{MedR $\downarrow$} \\
\midrule
(a) All 
& \textbf{\texttt{TEMOS}~\cite{petrovich2022temos}} &  2.12 &  4.09 &  5.87 &  8.26 & 13.52 & 173.0 &  3.86 &  4.54 &  6.94 &  9.38 & 14.00 & 183.25 \\
&        \textbf{HumanML3D~\cite{guo2022humanml3d}} &  1.80 &  3.42 &  4.79 &  7.12 & 12.47 & 81.00 &  2.92 &  3.74 &  6.00 &  8.36 & 12.95 & 81.50 \\

&                         \textbf{TMR~\cite{petrovich2023tmr}} &  5.68 & 10.59 & 14.04 & 20.34 & 30.94 & 28.00 &  \textbf{9.95} & 12.44 & 17.95 & 23.56 & 32.69 & 28.50 \\

&                         \textbf{Ours (distilbert)} & \underline{8.13} & \underline{14.16} & \underline{19.69} & \underline{27.07} & \underline{39.18} & \underline{18.00} &  \underline{9.29} & \underline{15.51} & \underline{20.61} & \underline{28.29} & \underline{40.25} & \underline{18.00} \\
&                         \textbf{Ours (RoBERTa)} &  \textbf{9.05} & \textbf{15.23} & \textbf{20.47} & \textbf{28.62} & \textbf{41.60} & \textbf{16.00} &  9.01 & \textbf{15.92} & \textbf{21.42} & \textbf{29.50} & \textbf{41.43} & \textbf{16.00} \\


             
\midrule
(b) All with threshold 
&         \textbf{\texttt{TEMOS}~\cite{petrovich2022temos}} &  5.21 &  8.22 & 11.14 & 15.09 & 22.12 & 79.00 &  5.48 &  6.19 &  9.00 & 12.01 & 17.10 & 129.0 \\
&        \textbf{HumanML3D~\cite{guo2022humanml3d}} &  5.30 &  7.83 & 10.75 & 14.59 & 22.51 & 54.00 &  4.95 &  5.68 &  8.93 & 11.64 & 16.94 & 69.50 \\
&        \textbf{TMR~\cite{petrovich2023tmr}} & \textbf{11.60} & 15.39 & 20.50 & 27.72 & 38.52 & \underline{19.00} & \textbf{13.20} & 15.73 & 22.03 & 27.65 & 37.63 & \underline{21.50} \\

&        \textbf{Ours (distilbert)} & 10.82 & \underline{18.49} & \underline{25.33} & \underline{33.89} & \underline{46.54} & \textbf{12.00} & \underline{12.25} & \textbf{19.69} & \underline{24.98} & \underline{32.70} & \underline{44.04} & \textbf{14.00} \\
&        \textbf{Ours (RoBERTa)} & \underline{11.39} & \textbf{19.18} & \textbf{25.73} & \textbf{34.76} & \textbf{47.94} & \textbf{12.00} & 11.65 & \underline{19.54} & \textbf{25.45} & \textbf{33.67} & \textbf{45.08} & \textbf{14.00} \\


                            
\midrule
(c) Dissimilar subset 
&        \textbf{\texttt{TEMOS}~\cite{petrovich2022temos}} & 33.00 & 42.00 & 49.00 & 57.00 & 66.00 &  4.00 & 35.00 & 44.00 & 50.00 & 56.00 & 70.00 &  3.50 \\
&        \textbf{HumanML3D~\cite{guo2022humanml3d}} & 34.00 & 48.00 & 57.00 & 72.00 & 84.00 &  3.00 & 34.00 & 47.00 & 59.00 & 72.00 & 83.00 &  3.00 \\

&                         \textbf{TMR~\cite{petrovich2023tmr}} & \underline{47.00} & 61.00 & \underline{71.00} & \underline{80.00} & 86.00 &  \underline{2.00} & \underline{48.00} & \underline{63.00} & 69.00 & 80.00 & 84.00 &  \underline{2.00}\\

&  \textbf{Ours (distilbert)} & 45.73 & \underline{62.80} & 70.73 & 79.88 & \textbf{90.85} & \underline{2.00} & 46.95 & 62.80 & \underline{70.12} & \underline{82.93} & \textbf{91.46} &  \underline{2.00}\\

&  \textbf{Ours (RoBERTa)} & \textbf{57.73} & \textbf{78.35} & \textbf{81.44} & \textbf{86.60} & \underline{90.72} &  \textbf{1.00} & \textbf{63.92} & \textbf{80.41} & \textbf{82.47} & \textbf{87.63} & \underline{90.72} &  \textbf{1.00} \\


\midrule
(d) Small batches \cite{guo2022humanml3d} 
& \textbf{\texttt{TEMOS}~\cite{petrovich2022temos}} & 40.49 & 53.52 & 61.14 & 70.96 & 84.15 &  2.33 & 39.96 & 53.49 & 61.79 & 72.40 & 85.89 &  2.33 \\
&        \textbf{HumanML3D~\cite{guo2022humanml3d}} & 52.48 & 71.05 & 80.65 & 89.66 & 96.58 &  1.39 & 52.00 & 71.21 & 81.11 & 89.87 & \underline{96.78} &  1.38 \\
&         \textbf{TMR~\cite{petrovich2023tmr}} & 67.16 & 81.32 & 86.81 & 91.43 & 95.36 &  \underline{1.04} & 67.97 & 81.20 & 86.35 & 91.70 & 95.27 &  \underline{1.03} \\

&         \textbf{Ours (distilbert)} & \underline{72.28} & \underline{85.42} & \textbf{90.15} & \textbf{94.01} & \textbf{97.09} &  \textbf{1.00} & \underline{72.21} & \underline{85.19} & \underline{90.00} & \textbf{94.42} & \textbf{97.04} &  \textbf{1.00} \\ 

&         \textbf{Ours (RoBERTa)} & \textbf{72.88} & \textbf{85.54} & \underline{89.91} & \underline{93.46} & \underline{96.68} &  \textbf{1.00} & \textbf{73.00} & \textbf{85.64} & \textbf{90.17} & \underline{93.70} & 96.49 &  \textbf{1.00} \\ 
\bottomrule        
\end{tabular}
}
\vspace{-0.5cm}
\end{table*}

\begin{table*}[ht]
\centering
\caption{\textbf{Comparison of text-to-motion generation on HumanML3D.} For each metric, we repeat the evaluation 20 times and report the average with 95$\%$ confidence interval. The right arrow (→) indicates that the closer the result is to real motion, the better.}
\vspace{-12pt}
\scalebox{0.95}{
\begin{tabular}{lccccccc} 
\hline
\multirow{2}{*}{Methods} & \multicolumn{3}{c}{R-Precision $\uparrow$}                                                                                                                                              & \multirow{2}{*}{FID $\downarrow$}                            & \multirow{2}{*}{MM-Dist $\downarrow$}                       & \multirow{2}{*}{Diversity $\rightarrow$}                    & \multirow{2}{*}{MModality $\uparrow$}        \\ 
\cline{2-4}
                         & Top-1 $\uparrow$                                            & Top-2 $\uparrow$                                            & Top-3 $\uparrow$                                            &                                                              &                                                             &                                                             &                                              \\ 
\toprule
Real & $0.511^{\pm .003}$ & $0.703^{\pm .003}$ & $0.797^{\pm .002}$ & $0.002^{\pm .000}$ & $2.974^{\pm .008}$ & $9.503^{\pm .065}$ & - \\ 
\toprule
MDM \cite{tevet2022humanmotiondiffusionmodelmdm} & $0.320^{\pm .005}$ & $0.498^{\pm .004}$ & $0.611^{\pm .007}$ & $0.544^{\pm .044}$ & $5.566^{\pm .027}$ & $9.559^{\pm .086}$ & $\mathbf{2.799^{\pm .072}}$ \\
GuidedMotion \cite{jin2024localactionguidedmotiondiffusion} & $0.503^{\pm .002}$ & $0.691^{\pm .002}$ & $0.788^{\pm .002}$ & $0.057^{\pm .006}$ & $3.040^{\pm .012}$ & $9.864^{\pm .077}$ & $2.473^{\pm .096}$ \\
KP \cite{liu2023bridging} & $0.496$ & - & - & $0.275$ & - & $9.975$ & $2.218$ \\
FG-MDM \cite{fgmdm} & $0.374^{\pm .003}$ & $0.582^{\pm .003}$ & $0.709^{\pm .005}$ & $0.618^{\pm .009}$ & $5.274^{\pm .048}$ & $9.563^{\pm .0.097}$ & - \\
FineMoGen \cite{finemogen} & $0.504^{\pm .002}$ & $0.690^{\pm .002}$ & $0.784^{\pm .002}$ & $0.151^{\pm .008}$ & $2.998^{\pm .008}$ & $9.263^{\pm .094}$ & \underline{$2.696^{\pm .079}$} \\
MotionLCM \cite{motionlcm} & $0.504^{\pm .002}$ & $0.698^{\pm .003}$ & $0.796^{\pm .002}$ & $0.304^{\pm .003}$& $3.012^{\pm .007}$ & $9.634^{\pm .064}$ & $2.267^{\pm .082}$\\
ParCo \cite{parco} & $0.515^{\pm .003}$ & $0.706^{\pm .003}$ & $0.801^{\pm .002}$ & $0.109^{\pm .005}$& $2.927^{\pm .008}$ & $9.576^{\pm .088}$ & $1.382^{\pm .060}$\\
MMM \cite{pinyoanuntapong2024mmm}& $0.504^{\pm .003}$ & $0.696^{\pm .003}$ & $0.794^{\pm .002}$ & $0.080^{\pm .003}$ & $2.998^{\pm .007}$ & $9.411^{\pm .058}$ & $1.164^{\pm .041}$ \\
MoMask \cite{guo2024momask} & $0.521^{\pm .002}$ & $0.713^{\pm .002}$ & $0.807^{\pm .002}$ & \underline{$0.045^{\pm .003}$} & $2.958^{\pm .008}$ & $9.678^{\pm .052}$ & $1.241^{\pm .040}$ \\
\toprule
Ours (CLIP) & ${\underline{0.529^{\pm .003}}}$ & ${\underline{0.722^{\pm .002}}}$ & $\underline{0.817^{\pm .002}}$ & $0.050^{\pm .003}$ & $\underline{2.907^{\pm .009}}$ & $9.684^{\pm .063}$ & $1.313^{\pm .041}$ \\

Ours (HTMA) & $\mathbf{0.532^{\pm .002}}$ & ${\mathbf{0.724^{\pm .003}}}$ & $\mathbf{0.821^{\pm .003}}$ & $\mathbf{0.039^{\pm .003}}$ & $\mathbf{2.901^{\pm .010}}$ & $9.674^{\pm .058}$ & $1.321^{\pm .039}$ \\
\bottomrule
\end{tabular}
}
\vspace{-0.6cm}
\label{tab:humanml3d}
\end{table*}

\begin{figure}[thb]
\centering
\includegraphics[width=\linewidth]{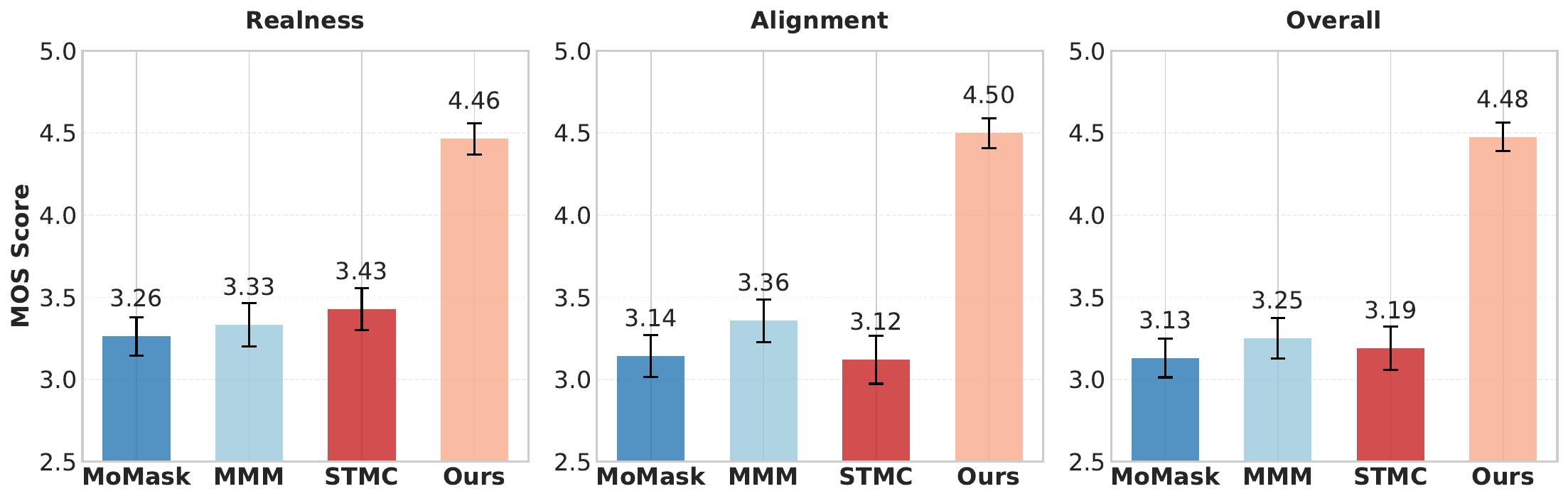}
\vspace{-0.8cm}
\caption{\textbf{User Study}. We generate 80 videos for each method to assess \textit{Realness}, \textit{T2M Alignment}, and \textit{Overall Impression}.}
 \vspace{-0.3cm}
\label{fig:user-study}
\end{figure}

We conduct experiments on the motion-text benchmark dataset, HumanML3D~\cite{guo2022humanml3d}, which collects 14,616 motions from AMASS~\cite{mahmood2019amass} and HumanAct12~\cite{guo2020action2motion} datasets, with each motion described by 3 text scripts, totaling 44,970 descriptions. We adopt their pose representation and augment the dataset using mirroring, followed by a 80/5/15 split for training, validation, and testing, akin to previous work~\cite{pinyoanuntapong2024mmm,guo2024momask}. Our KinMo dataset is built on this dataset with each motion described by 6 group-level and 15 interaction-level descriptions scripts in accordance with human kinematics (see Sec.~\ref{sec:dataset}). An example is presented in the supplementary material (Appendix \ref{sec:sub_dataset}), along with additional details on dataset collection. 
All experiments are performed in such settings, as shown in \cref{fig:motiongeneration}.

\subsection{Text-Motion Retrieval}
We first evaluate whether the introduction of group- and interaction-level motion descriptions reduces any ambiguity for the text-motion retrieval problem and improves the overall motion understanding.

\noindent\textbf{Evaluation Metrics.} 
We adopt TMR settings~\cite{petrovich2023tmr} to measure retrieval performance using recall scores at various ranks ({\eg}, R@1, R@2) and the median rank (MedR) of our results. MedR represents the median ranking position of the ground-truth result, with lower values indicating more precise retrievals. The four evaluation protocols used in our experiments are outlined below: (i) \textit{All} uses the complete test dataset, though similar negative pairs can affect precision; (ii) \textit{All with threshold} sets a similarity threshold of 0.8 to determine accurate retrievals; (iii) \textit{Dissimilar subset} uses 100 distinctly different sampled pairs measured by sBERT~\cite{reimers2019sentence} embedding difference; and (iv) \textit{Small batches} evaluates performance on random batches of 32 motion-text pairs.

\begin{figure*}[thb]
\centering
\includegraphics[width=0.95\linewidth]{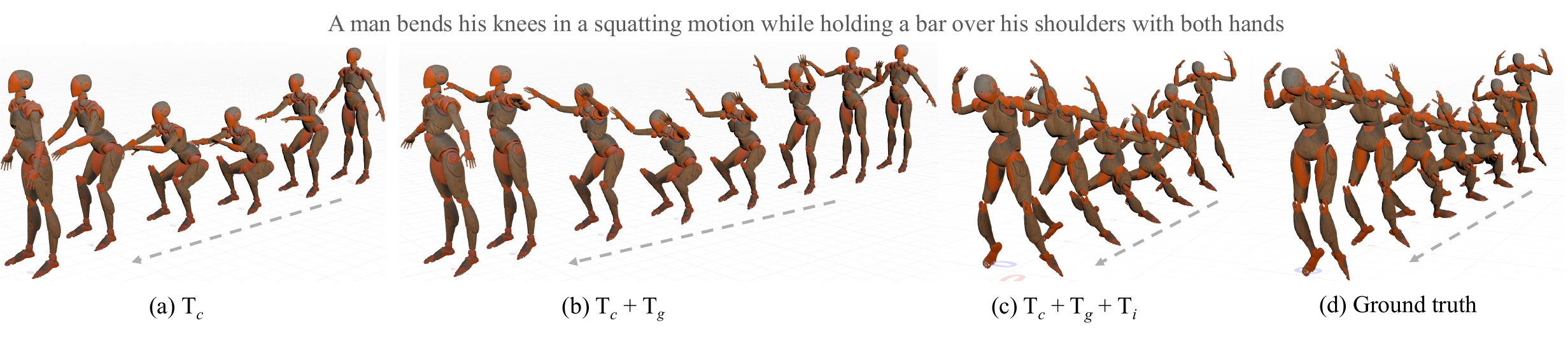}
\vspace{-0.55cm}
\caption{\textbf{Additional Descriptions resolve the ambiguity of motion generation}. We create motion using (a) only global action, (b) global + group-level descriptions, and (c) global, group-level, and interaction-level descriptions, and compare them with (d) the ground truth.}
 \vspace{-0.3cm}
\label{fig:ablation-visual-gen}
\end{figure*}

\noindent\textbf{Evaluation Results.}
We benchmark KinMo against \cite{petrovich2022temos, guo2022humanml3d, petrovich2023tmr}. In \cref{tab:benchmark_h3d}, our model outperforms existing baselines, particularly in setting (a). This improvement is primarily due to our annotated descriptions, which help to resolve ambiguities in action-level text-motion correspondence. By providing finer-grained details, our approach enhances the discrimination of motions with subtle local movement differences but similar global action descriptions. Our proposed formulation and annotations contribute significantly to motion understanding by capturing intricate local movement details throughout the motion sequence.
Motion understanding can be further enhanced using RoBERTa~\cite{roberta} as a stronger text encoder for additional descriptions.


\subsection{Text-Motion Generation}
\label{sec:exp_t2m}
\noindent\textbf{Evaluation Metrics.} We adopt (1) \textit{FID}~\cite{heusel2017fid} as an overall motion quality metric to measure the difference between generated and real motion distributions; (2) \textit{R-Precision (R-Prec)} and \textit{multimodal distance (MM-Dist)} to quantify the semantic alignment between text and generated motions; and (3) \textit{Multimodality (MModality)} to assess the diversity of motions generated from the same text, as in T2M~\cite{guo2022humanml3d}\looseness=-1. 

\noindent\textbf{Evaluation Settings.} To present a fair comparison, we consider each T2M method as the whole system. 
For KinMo, we apply a different random seed during inference. Motion Reasoner generates the group- and interaction-level motion descriptions based on the provided global action descriptions and feeds them into the generator.

\noindent\textbf{Evaluation Results.} 
\cref{tab:humanml3d} compares KinMo with various methods for T2M generation~\cite{tevet2022humanmotiondiffusionmodelmdm, zhang2022motiondiffuse, guo2024momask, pinyoanuntapong2024mmm, petrovich2022temos, motionlcm,fgmdm,jin2024localactionguidedmotiondiffusion,finemogen,parco,liu2023bridging}. Our method attains the best motion generation quality with the highest text alignment score (R-Prec and MM-Dist). Thanks to the introduction of explicit group- and interaction-level descriptions, we observe that KinMo generates better aligned motions for any given dense and fine-grained text descriptions shown in ~\cref{fig:visual_comparison}, while other baseline methods fail to capture local body part movements.

\noindent\textbf{User Study.} 
To assess the quality of our results, we conduct a user study involving 20 participants and 320 samples, 80 from KinMo, MoMask~\cite{guo2024momask}, MMM~\cite{pinyoanuntapong2024mmm}, and STMC~\cite{petrovich24stmc}, respectively. Each participant was presented the video clips in a random order and asked to rate the results between 1 (lowest) and 5 (highest) based on (1) \textit{realness}, (2) correctness of text-motion \textit{alignment}, and (3) \textit{overall impression}. \cref{fig:user-study} shows that, unlike other baseline methods, KinMo achieves higher Mean Opinion Scores (MOS) overall.

\begin{table}
\renewcommand{\baselinestretch}{0.5}
\caption{\textbf{Comparisons with other methods}. Motion Generation with CLIP-embed additional texts as conditions of MoMask.}
\label{tab:ab-1}
\vspace{-10pt}
\centering
\setlength{\tabcolsep}{4pt}
\resizebox{\linewidth}{!}{
\begin{tabular}{c|cccc}
\toprule
Method & FID$\downarrow$ & R-Prec(Top 3)$\uparrow$ & MM-Dist $\downarrow$ & MModality $\uparrow$ \\
\midrule
MoMask(base) & \textbf{0.045} & 0.807 & 2.958 & 1.241 \\
MoMask+LGTM  & 0.057  & 0.801 & 2.963  & 1.123\\
MoMask+FinMoGen& 0.062 & 0.799  & 2.998 & 1.223\\
\midrule
Ours+Parco (2-group) & 0.077  & 0.793 & 3.232  & 1.101\\
Ours (MoMask+KinMo) (6-group) & 0.050 & \textbf{0.817}  & \textbf{2.907} & \textbf{1.313}\\
\bottomrule
\end{tabular}
}
\end{table}


\begin{table}
\caption{\textbf{Effect of Additional Descriptions for Text-Motion Alignment.} Different strategies for incorporating descriptions generated by Motion Reasoner on motion- and text-retrieval tasks.}
\label{tab:ab-alignment}
\vspace{-10pt}
\centering
\setlength{\tabcolsep}{4pt}
\resizebox{0.99\linewidth}{!}{
\begin{tabular}{c|cccc|cccc}
    \toprule
        \textbf{Motion} &  \multicolumn{4}{c|}{Text-motion retrieval} & \multicolumn{4}{c}{Motion-text retrieval} \\
       \textbf{Semantic} & \small{R@1 $\uparrow$} & \small{R@2 $\uparrow$} & \small{R@3 $\uparrow$}  & \small{MedR $\downarrow$} & \small{R@1 $\uparrow$} & \small{R@2 $\uparrow$} & \small{R@3 $\uparrow$} & \small{MedR $\downarrow$} \\
\midrule
global & 3.67 & 7.17 & 10.32 & 40.00 & 8.08 & 11.56 & 17.23 & 38.00  \\
+ group & 7.58 & 13.16 & 16.97 & 22.00 & 8.58 & 14.51 & 19.21  & 21.00  \\
+ interact & \textbf{9.05} & \textbf{15.23} & \textbf{20.47} &  \textbf{16.00} & \textbf{9.01} & \textbf{15.92} & \textbf{21.42} &  \textbf{16.00} \\
- cross &  7.63 & 13.13 & 16.94 & 22.00 &  8.60 & 14.54 & 19.21 & 21.00 \\
\bottomrule

\end{tabular}
}
\vspace{-0.2cm}
\end{table}



\begin{table}
\caption{\textbf{Effect of Hierarchical Text-Motion Alignment.} Comparisons are conducted for Motion Generator with RQ base layer.}
\label{tab:ab-gen}
\vspace{-10pt}
\centering
\setlength{\tabcolsep}{4pt}
\resizebox{0.99\linewidth}{!}{
\begin{tabular}{c|ccccccc}
\toprule
\multirow{1}{*}{Embedder} & \multirow{1}{*}{Global} & \multirow{1}{*}{Joint} & \multirow{1}{*}{Inter} & FID$\downarrow$ & R-Prec(Top 3)$\uparrow$ & MM-Dist $\downarrow$ & MModality $\uparrow$ \\

\midrule
\multirow{3}{*}{CLIP} & \ding{51}& --  & --   & 0.115  & 0.499 & 2.999 & 1.221\\
& \ding{51}& \ding{51}  & -- & 0.096 & 0.503  & 2.953 & \textbf{1.308}\\
&  \ding{51}& \ding{51}& \ding{51} & 0.098 & 0.512  & 2.912 & \textbf{1.308}\\

\midrule
\multirow{3}{*}{HTMA} & \ding{51}& --  & --   & 0.056  & 0.512 & 2.969  & 1.232\\
& \ding{51}& \ding{51}  & -- & 0.051 & 0.525  & 2.911 & 1.292\\
&  \ding{51}& \ding{51}& \ding{51} & \textbf{0.044} & \textbf{0.527}  & \textbf{2.904} & 1.305\\
\bottomrule
\end{tabular}
}
\vspace{-0.4cm}
\end{table}

\subsection{Ablation Study}

\noindent \textbf{Effect of Additional Descriptions generated by Motion Reasoner at each level.}
\cref{tab:ab-alignment} summarizes several strategies for incorporating text-motion alignment: (1) only global action (global), (2) + group-level (+ group), (3) + group + interaction-level (+ interact), and (4) without cross-attention (- cross). We observe that adding extra descriptions generated by Motion Reasoner enhances motion understanding. Cross-attention improves the connectivity of descriptions from different hierarchy levels. As shown in \cref{fig:ablation-visual-gen}, both group- and interaction-level descriptions are beneficial for resolving global action ambiguity and generating local body parts (e.g., the hands and arms in the figure). Refer to Appendix \ref{sec:sub_addn_expts} for further analysis of the order of the different descriptions and additional design choices.\looseness=-1

\noindent \textbf{Effect of Hierarchical Text-Motion Alignment (HTMA).}
We provide additional quantitative results and comparisons for various text encoders. As demonstrated in \cref{tab:ab-gen}, the CLIP encoder, as used in previous work~\cite{guo2024momask,pinyoanuntapong2024mmm}, shows superior text-motion alignment after complete training. Our HTMA method enhances motion smoothness and naturalness, as evident by a significantly lower FID. For the motion generation procedure, it can be seen that both of these text encoders benefit from our coarse-to-fine generation approach. Further analysis of training is given in Appendix \ref{sec:sub_addn_expts}.\looseness=-1

\noindent \textbf{Text Granularity and Motion Decomposition.} 
Several methods, including LGTM~\cite{lgtm}, FG-MDM~\cite{fgmdm}, and FinMoGen~\cite{finemogen}, employ LLMs to generate supplementary motion descriptions to improve generation. KinMo formulates the generated supplementary descriptions using insight of motion components (position, angle, velocity) with natural language to enhance text-motion alignment. We validate this advantage via quantitative experiments (Tab.~\ref{tab:ab-1}) and ensure fairness using MoMask as the generator across all methods, with only additional descriptions replaced.
Moreover, we compare with ParCo~\cite{parco} which decomposes motion into 2 parts (upper and lower body) as opposed to our proposed 6 parts based on kinematic knowledge. KinMo outperforms these approaches, indicating that (1) the formulated text descriptions, instead of random ones, improve model performance and (2) the proposed linguistically describable motion representation based on kinematic parts (Sec~\ref{sec:representation}) and corresponding descriptions are necessary.

\subsection{Applications}
\label{sec:exp-applications}
\noindent\textbf{Text-to-Motion Editing.} 
\textit{Motion Reasoner} enables precise action-level edits (e.g., changing \textit{running} to \textit{jumping}) or local joint adjustments (e.g., \textit{slightly raising the hands}). Our method uses a coarse-to-fine approach, assisted by a masking mechanism, to perform these edits at varying levels of granularity. Please refer to Appendix \ref{sec:sub_addn_expts} for evaluation.

\noindent\textbf{Motion Trajectory Control.} We employ ControlNet~\cite{omnicontrol} to condition the motion generator using the provided trajectory of the target joint during the generation, with the descriptions adjusted by the \textit{Motion Reasoner}. We defer the technical details to Appendix \ref{sec:sub_implement}.


%% file: sec/5_conclusion.tex
\vspace{-0.2cm}
\section{Conclusion}
\label{Conclusion}
\vspace{-0.1cm}

We present \textbf{KinMo}, a framework that represents human motion as kinematic parts movements and interactions, thereby enabling fine-grained text-to-motion understanding, generation, editability, and control. Our method progressively encodes global actions with kinematic descriptions and leverages these descriptions to achieve enhanced alignment and understanding, thus generating coarse-to-fine motions. The KinMo dataset is publicly available to the scientific community. Extensive comparisons with state-of-the-art methods show that \textbf{KinMo} improves text-motion alignment and body part control. 

\noindent \textbf{Acknowledgements.} 
We thank the anonymous reviewers for their constructive feedback. Special thanks to Brian Burritt and Avi Goyal for helping with visualizations, and Kyle Olszewski and Ari Shapiro for valuable discussions.

%% file: arxiv_sec/supp.tex
\clearpage
\appendix

\begin{center} 
    \centering
    \textbf{\large KinMo: Kinematic-aware Human Motion Understanding and Generation}
\end{center}
\begin{center} 
    \centering
    \large Supplementary Material 
\end{center}

\section{Overview}
\label{sec:Summary}
This supplementary document contains further details on dataset collection and our framework, additional ablation studies and experimental results, and limitations of KinMo. For more visual results, refer to the demo videos on the project page: {\small\url{https://andypinxinliu.github.io/KinMo}}.
    

\section{Dataset Collection Details}
\label{sec:sub_dataset}
Unlike conventional dataset annotation approaches, which require extensive human labeling and verification, our semi-automatic data collection pipeline with human in the loop significantly reduces the cost while providing high-quality annotation. The procedure is described in the following: 

\vspace{0.1cm}
\noindent \textbf{Pose Text Description Generation.} To generate text descriptions for individual poses, we adopt PoseScript~\cite{delmas2022posescript}, which provides detailed descriptions for each pose by capturing fine-grained details of joint positions and their spatial relationships. In addition to providing a textual representation of the pose, PoseScript is capable of describing the relative positions of different joints, which is crucial for understanding complex human motion. 

\vspace{0.1cm}
\noindent \textbf{Keyframe Selection.} While PoseScript offers per-frame annotations, it does not provide a direct way to capture the temporal transitions between poses over time. 
We observe that text descriptions for temporally adjacent frames are often very similar. In contrast, frames that are farther apart in time exhibit less overlap in their descriptions, often presenting different semantics.

\begin{figure}[t]
\centering
\includegraphics[width=\linewidth]{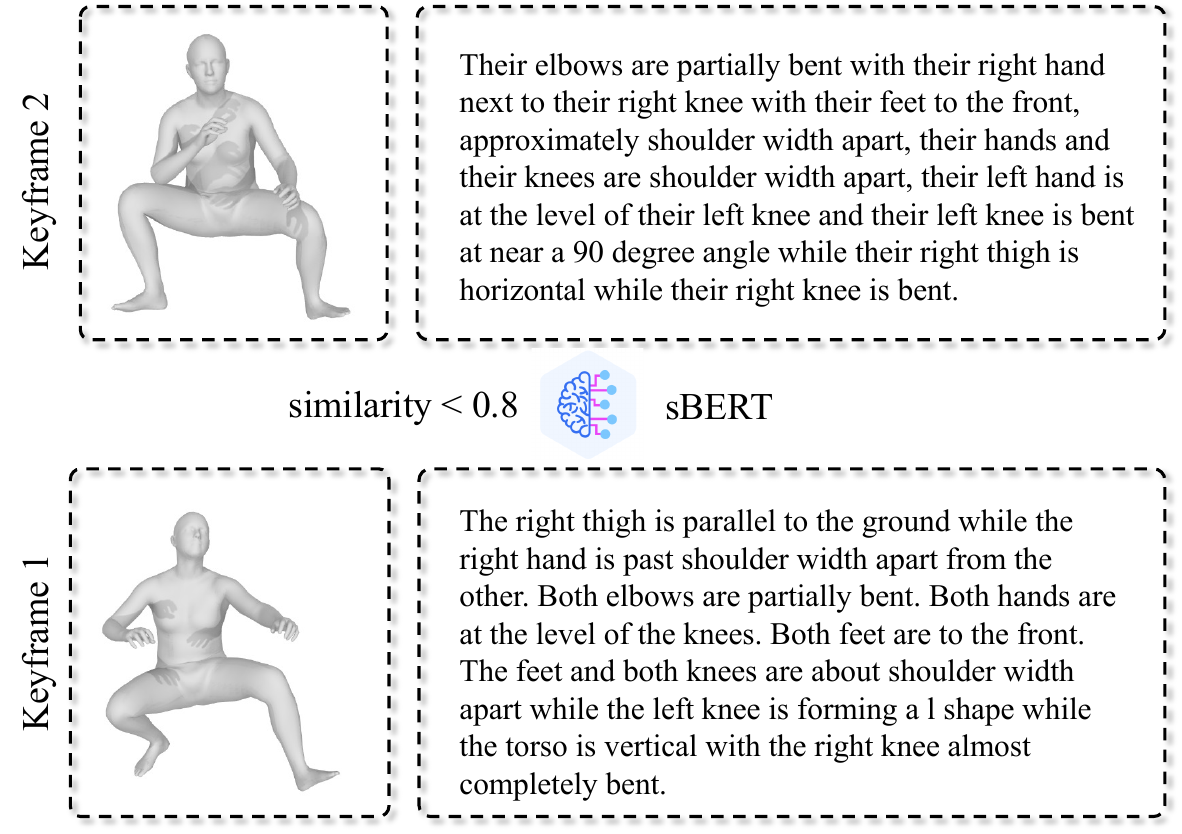}
\vspace{-2em}
\caption{\textbf{Visualization of Sample Keyframes.} We use PoseScript to obtain per-frame text annotations for poses and select keyframes based on sentence similarities.}
\vspace{-1em}
\label{fig:threshold}
\end{figure}

Based on this observation, we devise a keyframe-based approach detailed as follows. We utilize sBERT~\cite{reimers2019sentence} to extract embeddings for the PoseScript-generated descriptions of each frame. By calculating the cosine similarity between these text embeddings, we can measure the similarity of poses across frames. If the cosine similarity between two frames falls below a threshold of 0.8, we classify the frame as a keyframe, marking a significant temporal transition. This allows us to isolate key moments in the sequence that represent meaningful pose changes and filter out redundant frames for subsequent analysis. We then compute temporal local motions by analyzing kinematic group differences across a specified time window, allowing us to capture finer motion details within the pose sequence. \cref{fig:threshold} shows a visualization of sample keyframes within our dataset.

\begin{table}[t]
\centering 
\small
\caption{\textbf{Prompt template for automatic dataset annotation}. We show one specific example in \cref{tab:example} with more details.}
\label{tab:template}
\vspace{-0.3cm}
\begin{tcolorbox}

\textbf{System Prompt}
\newline
You are a motion description labeler who should describe the motion using your language as detailed as possible. Now, describe the motion in the given video or text by describing the motion of each kinematic group respectively: Kinematic groups: torso, neck, left arm, right arm, left leg, right leg.
\newline
\newline
\textcolor[rgb]{1,0,0}{[Answer Format Instruction]}
\newline
To describe a motion, you should describe Inside each group and Between groups:
\newline
INDIVIDUAL GROUP: torso: \textless MOTION DESCRIPTION\textgreater ; neck: \textless MOTION DESCRIPTION\textgreater ...
\newline
\newline
\textcolor[rgb]{1,0,0}{[Answer Content Instruction]}
\newline
In your description, you can use simple adjectives or numerical scale (distance/degree/speed) to describe each motion (for example, push forward for 3 meters, from 0 to 45 degrees, etc)...
\newline
\newline
\textbf{User Prompt}
\newline
\textcolor[rgb]{0,0,0.9}{[One-shot Example for In-Context-Learning]}
\newline
Think about the motion: [a man stumbles to his right. the motion seems sudden so he was probably pushed. a person standing loses balance, falling to the right and recovers standing...
\newline
\newline
\textcolor[rgb]{0,0.9,0}{[Data to be Annotated]}
\end{tcolorbox}
\vspace{-0.3cm}

\end{table}

\vspace{0.1cm}
\noindent \textbf{Temporal Joint Motion Reasoning.} For reasoning about the potential motions of kinematic groups across keyframes, we leverage a Large Language Model (LLM), specifically GPT-4~\cite{openai2024gpt4technicalreport}. Given the text descriptions of the selected keyframes, GPT-4 is tasked with generating potential motions for each kinematic group by reasoning through the temporal relationships between poses. To guide the LLM, we provide a well-structured prompt design, shown in \cref{tab:template}. We give a system prompt to use GPT-4 as a motion labeler, providing a specific answer format instruction and content instruction. During each query, the user prompt contains one specific annotation example with explanation for one-shot in-context-learning and the keyframe descriptions needed to be annotated. A real example prompt for a single motion sequence is provided in \cref{tab:example}, and its corresponding annotation result is shown in \cref{tab:sample}. This design allows the model to infer the most likely motions of specific kinematic groups and predict how they may evolve over time. The reasoning process enables the system to generate coherent and realistic motion sequences based on the selected keyframes.

\vspace{0.1cm}
\noindent \textbf{Human Evaluation.} Human evaluators play a critical role in ensuring the accuracy and quality of text and motion predictions generated by the system. The evaluation process is conducted in two stages:
\begin{itemize}
    \item \textbf{Keyframe Selection Evaluation.} Two human evaluators review the selected keyframes by the system and the corresponding rendered motion sequences. They identify the optimal cosine similarity threshold for filtering keyframes and examine whether the selected keyframes adequately capture the most significant pose transitions. The threshold is iteratively adjusted based on the evaluators' feedback to improve the precision of keyframe selection.
    \item \textbf{Text Description Evaluation.} The evaluators also assess the generated text descriptions for each joint-level motion. They determine whether the descriptions accurately reflect the motion dynamics and satisfy the intended criteria. If errors or inconsistencies are found, the evaluators provide feedback to the system, prompting a revision of the current prompt design. This iterative process continues until the evaluators reach a consensus, measured by a Cohen’s Kappa statistic of at least 0.8, indicating a strong agreement.
\end{itemize}
The evaluators spent about one day interacting with LLM for the iterative prompt optimization. Both the average prompt length and returning answer is less than 1000 words, resulting in approximately 3200 tokens for both input and output, and \textbf{23 USD} final cost for the whole annotation (44,970 motion sequences).

\vspace{0.1cm}
\noindent \textbf{Accuracy of KinMo Annotation.} We conducted two complementary evaluations: (a) \textit{Human evaluation:} Five independent annotators (not involved in the annotation) assessed 500 randomly selected samples. Each annotator viewed the motion video and scored the associated description on a scale of 0–10 for three criteria: spatial accuracy, temporal coherence, and consistency with the global text. The average scores in \cref{tab:reb_1} indicate strong alignment between motion and text. (b) \textit{LLM-based evaluation:} We used GPT-4o-mini~\cite{openai2024gpt4technicalreport} to perform the same evaluation, using a structured prompt (\cref{tab:template}) and scoring based on the same criteria. LLM-based scores were similarly high, with $\leq$5\% of samples flagged as potentially inconsistent. These evaluations demonstrate that KinMo’s auto-generated descriptions reliably capture both spatial and temporal motion details at scale, reinforcing the quality of our dataset beyond indirect task-based metrics.

\begin{table}[t]
\centering
\caption{\textbf{Evaluation Results of KinMo detailed descriptions}. Bad Response Rate (BRR) is assessed by $BRR=\frac{items~with~score < 5}{total~items}$.}
\vspace{-0.35cm}
\label{tab:reb_1}
\resizebox{0.9\linewidth}{!}{
\begin{tabular}{l|ccc|c}
\toprule
Evaluation & preservation of & capture of & consistency with & Metrics \\
Method & spatial detail & temporal dynamics & the global text &  \\
\midrule
Human & 8.24 & 8.01 & 8.71 & Average Score  \\
evaluation & 2.96\% & 3.15\% & 1.32\% & BRR \\
             
\midrule
LLM & 8.30 & 8.12 & 8.55 & Average Score  \\
evaluation & 2.68\% & 2.37\% & 1.55\% & BRR \\

\bottomrule        
\end{tabular}
}
\end{table}

\begin{table*}
    \centering
    \small
    \setlength{\tabcolsep}{4pt}
    \caption{\small{\textbf{Global Action Text, Low-level Text and Motion as Tri-modality Retrieval Benchmark on HumanML3D~\cite{guo2022humanml3d}.} HText denotes Global Action-level descriptions, LText demotes joint-level motion descriptions.
    }}
    \label{tab:tri-alignment}
    \resizebox{0.99\linewidth}{!}{
    \begin{tabular}{l|cccccc|cccccc}
        \toprule
        
         \multirow{2}{*}{\textbf{Setting}}  & \multicolumn{6}{c|}{HText-Motion Retrieval} & \multicolumn{6}{c}{Motion-HText Retrieval} \\
         
         & \small{R@1 $\uparrow$} & \small{R@2 $\uparrow$} & \small{R@3 $\uparrow$} &  \small{R@5 $\uparrow$} & \small{R@10 $\uparrow$} & \small{MedR $\downarrow$} & \small{R@1 $\uparrow$} & \small{R@2 $\uparrow$} & \small{R@3 $\uparrow$} & \small{R@5 $\uparrow$} & \small{R@10 $\uparrow$} & \small{MedR $\downarrow$}\\
        \midrule
    
        (a) All  & 3.67 & 7.17 & 10.32 & 15.73 & 25.12 & 40.00 & 4.39 & 8.08 & 11.56 & 17.23 & 26.81 & 38.00  \\
        \midrule

        (b) All with threshold & 7.98 & 13.87 & 18.47 & 25.86 & 36.39 & 22.00 & 7.60 & 12.27 & 16.40 & 21.97 & 31.36 & 30.00 \\
        
        \midrule
        
        (c) Dissimilar subset & 34.15 & 52.44 & 58.54 & 72.56 & 81.10 & 2.00 & 37.20 & 54.88 & 62.80 & 68.90 & 79.27 & 2.00 \\
        \midrule
        (d) Small batches & 60.76 & 75.79 & 81.35 & 86.93 & 91.79 & 1.10 & 61.26 & 76.26 & 81.97 & 87.24 & 91.63 & 1.11 \\

        \midrule

        \multirow{2}{*}{\textbf{Setting}}  & \multicolumn{6}{c|}{LText-Motion Retrieval} & \multicolumn{6}{c}{Motion-LText Retrieval} \\
         
         & \small{R@1 $\uparrow$} & \small{R@2 $\uparrow$} & \small{R@3 $\uparrow$} &  \small{R@5 $\uparrow$} & \small{R@10 $\uparrow$} & \small{MedR $\downarrow$} & \small{R@1 $\uparrow$} & \small{R@2 $\uparrow$} & \small{R@3 $\uparrow$} & \small{R@5 $\uparrow$} & \small{R@10 $\uparrow$} & \small{MedR $\downarrow$}\\
        \midrule
    
        (a) All  & 3.57 & 7.17 & 9.82 & 14.77 & 24.30 & 37.00 & 4.15 & 8.17 & 11.39 & 15.89 & 25.42 & 37.00  \\
        \midrule

        (b) All with threshold & 7.12 & 12.39 & 16.65 & 23.85 & 35.48 & 22.00 & 7.31 & 12.06 & 16.06 & 21.09 & 31.21 & 30.00 \\
        
        \midrule
        
        (c) Dissimilar subset & 45.36 & 70.10 & 75.26 & 80.41 & 85.57 & 2.00 & 46.39 & 68.04 & 75.26 & 81.44 & 86.60 & 2.00 \\
        \midrule
        (d) Small batches & 62.21 & 78.29 & 83.97 & 88.57 & 93.08 & 1.08 & 63.10 & 77.98 & 83.87 & 88.74 & 93.15 & 1.05 \\

        \midrule

        \multirow{2}{*}{\textbf{Setting}}  & \multicolumn{6}{c|}{HText-LText Retrieval} & \multicolumn{6}{c}{LText-HText Retrieval} \\
         
         & \small{R@1 $\uparrow$} & \small{R@2 $\uparrow$} & \small{R@3 $\uparrow$} &  \small{R@5 $\uparrow$} & \small{R@10 $\uparrow$} & \small{MedR $\downarrow$} & \small{R@1 $\uparrow$} & \small{R@2 $\uparrow$} & \small{R@3 $\uparrow$} & \small{R@5 $\uparrow$} & \small{R@10 $\uparrow$} & \small{MedR $\downarrow$}\\
        \midrule
    
        (a) All  & 0.05 & 0.10 & 0.12 & 0.17 & 0.38 & 1905.0 & 1.72 & 3.00 & 4.34 & 6.41 & 10.89 & 194.0  \\
        \midrule

        (b) All with threshold & 0.07 & 0.12 & 0.17 & 0.57 & 1.12 & 1544.0 & 3.19 & 5.50 & 7.48 & 10.60 & 16.42 & 132.0 \\
        
        \midrule
        
        (c) Dissimilar subset & 1.03 & 2.06 & 4.12 & 6.19 & 15.46 & 48.00 & 22.68 & 36.08 & 45.36 & 55.67 & 72.16 & 4.00 \\
        \midrule
        (d) Small batches & 4.34 & 7.82 & 11.78 & 19.44 & 37.05 & 14.77 & 40.51 & 55.56 & 64.69 & 76.10 & 89.07 & 2.14 \\
                                      
    \bottomrule        
    \end{tabular}

    }
    \vspace{0.05in}
    
\end{table*}

\section{Additional Implementation Details}
\label{sec:sub_implement}

\noindent \textbf{Hierarchical Text-Motion Alignment.}
Both the motion and text encoders are based on Transformer architectures~\cite{transformer}. We add two tokens in front of the raw sequence to represent the mean and standard deviation, akin to those in the VAE-based ACTOR model~\cite{petrovich2022temos}. These encoders are probabilistic, generating parameters of a Gaussian distribution ($\mu$ and $\Sigma$) from which a latent vector $z \in \mathbb{R}^{d}$ can be sampled. The text encoder processes input features from a pretrained and frozen RoBERTa~\cite{roberta} model, while the motion sequence is given directly as input to the motion encoder. As for the cross-attention that connects three levels of semantics, we use one transformer block containing one multi-head attention layer with one MLP layer. For the neural network, we set the latent dimension to 512, the number of heads to 6, and the feed-forward size to 1024. We train this module for 70 epochs, with other settings the same as in TMR~\cite{petrovich2023tmr}.

\begin{figure*}
\centering
\includegraphics[width=\linewidth]{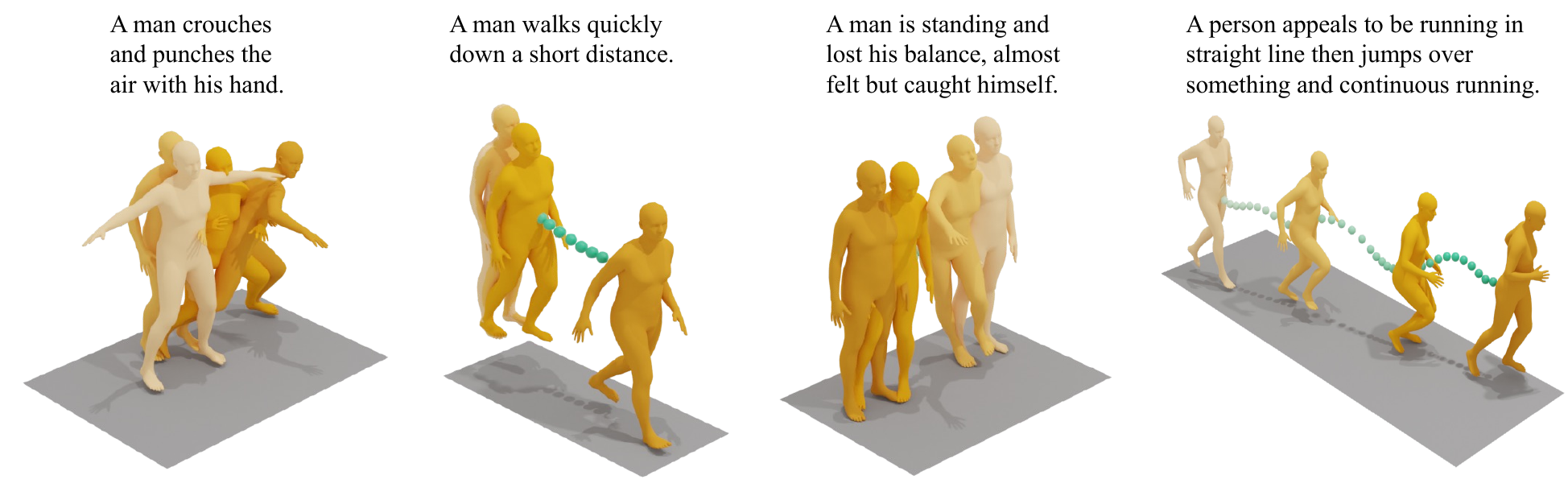}
\vspace{-2em}
\caption{\textbf{Visualization of Motion Trajectory Control}. We leverage pelvis locations to guide the motion generation in addition to the text.}
\label{fig:traj-control}
\end{figure*}

\vspace{0.1cm} 
\noindent \textbf{Text-Motion Generation.} We use MoMask \citep{guo2024momask} as our generator architecture. During the training process, to enhance the model's robustness to variations in text input, we randomly omit 10\% of the text conditioning. This approach also facilitates the use of Classifier-Free Guidance (CFG). Our codebook consists of 512 entries, with each having a 512-dimensional embedding and 6 residual layers. The Transformer’s embedding size is 384 and has 6 attention heads, each with an embedding dimension of 64, spread across 8 layers. Both the encoder and decoder reduce the motion sequence length by a factor of 4 when transitioning to the token space. The learning rate follows a linear warm-up schedule, peaking at 2e-4 after 2000 iterations. We utilize AdamW optimizer. The mini-batch size is 512 during the training of RVQ-VAE and 64 for training the Transformers. At inference, the CFG scale is set to $cfg = 4$ for the base layer and $cfg = 5$ for the 6 residual layers, with the generation process running for 10 iterations. To produce text embeddings, we apply Hierarchical Text-Motion Alignment (HTMA), which results in embeddings of size 512. These embeddings are subsequently reprojected to a 384-dimensional space to match the Transformer's token size.

\vspace{0.1cm}
\noindent \textbf{Motion Editing.}
We leverage a \textit{Joint Motion Reasoner} to refine both global and local action descriptions using the users' input. This model enables precise action-level edits (e.g., changing \textit{running} to \textit{jumping}) or local joint adjustments (e.g., \textit{slightly raising the hands}). Our method follows a coarse-to-fine approach, assisted by a masking mechanism, to perform these edits at varying levels of granularity. Specifically, by masking the target sequences and using the mask generator to fill in the masked area, we can dynamically adjust the motion to meet the target requirement. For more details on the masking-based editing process, please refer to MMM~\cite{pinyoanuntapong2024mmm}.

\begin{figure}
\centering
\includegraphics[width=\linewidth]{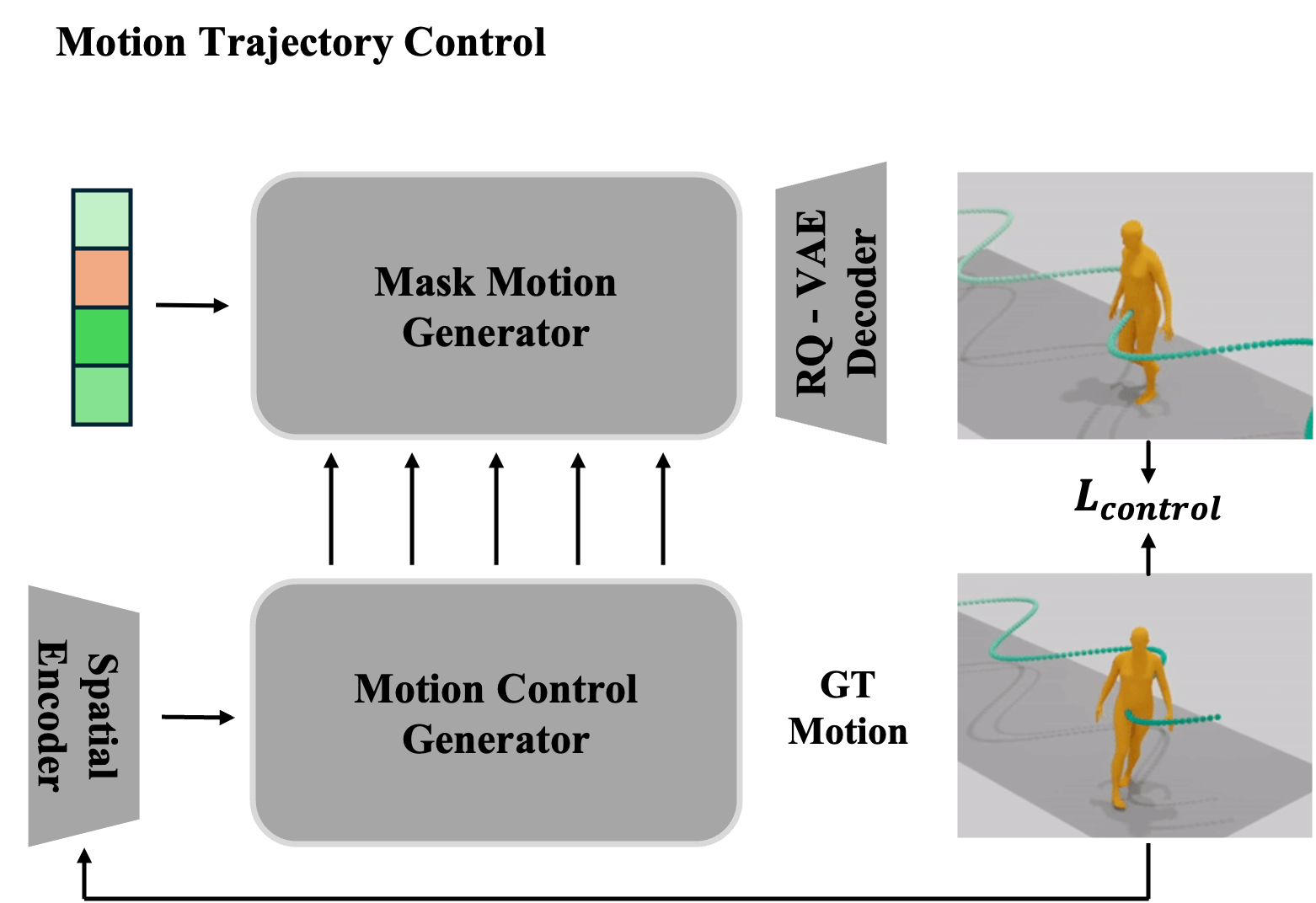}
\vspace{-2em}
\caption{\textbf{Motion Trajectory Control.} We adopt a ControlNet architecture to condition the generator with the provided trajectory of the target joint during the generation. We utilize a CNN encoder to process the spatial position information and feed it as the input condition into the control generator network.}
\vspace{-1em}
\label{fig:mo-control}
\end{figure}

\vspace{0.1cm}
\noindent \textbf{Motion Trajectory Control.}
Inspired by ControlNet~\cite{omnicontrol} for Diffusion Models, we incorporate the joint spatial conditioning for trajectory control. As shown in Fig.\ref{fig:mo-control}, during this stage, the motion control model is a trainable replica of the frozen mask motion generator. Specifically, each layer in the motion Control Generator is appended with a zero-initialized linear layer to remove random noise in the initial training steps. The initial $\tau$ poses are defined by the trajectories of $K$ control joints, $\mathbf{g}^{1:\tau} = \{\mathbf{g}^i\}_{i=1}^{\tau}$, where $\mathbf{g}^i \in \mathbb{R}^{K \times 3}$ denotes the global absolute locations of each control joint. A Trajectory Encoder ${\bf \Theta}^{b}$ consisting of convolution layers is used to encode the trajectory signals. Unlike previous methods like OmniControl~\cite{omnicontrol, tlcontrol}, which directly diffuses in the motion space to allow for explicit supervision of control signals, effectively supervising control signals in the latent space is non-trivial.
Therefore, in addition to using a motion reconstruction loss based on RQ-Tokenizer decoder ${\cal D}$ to decode the latent $\hat{\mathbf{z}}_{0}$ into the motion space, we also add a control loss ${\cal L}_\text{control}$ to obtain the predicted motion $\hat{\mathbf{x}}_0$:
\begin{equation}
    {\cal L}_\text{control} = \mathbb{E} \left[ \frac{\sum_{i} \sum_{j} m_{ij} || R(\hat{\mathbf{x}}_{0})_{ij} - R(\mathbf{x}_{0})_{ij}||_{2}^{2}}{\sum_{i} \sum_{j} m_{ij}} \right]\text{,}
    \label{eq:control_loss}
\end{equation}
where $R(\cdot)$ converts the joint local positions to global absolute locations and $m_{ij} \in \{0, 1\}$ is the binary joint mask at frame $i$ for the joint $j$.

In our network design, \textit{Motion Control Generator} is a trainable copy of \textit{Masked Transformer} with the zero linear layer connected to the output of each Masked Transformer layer in MoMask~\cite{guo2024momask} to mitigate random noise in the initial training steps. The spatial encoder is a 3-layer CNN-based Residual network with a temporal downsampling of factor 4 to encode the trajectory control signal (the joint position information to be controlled along the sequence).

\section{Additional Experimental Results}
\label{sec:sub_addn_expts}

\subsection{Text-Motion Alignment}
\label{sec:joint-analysis}

In this work, beyond the hierarchical text-semantics framework proposed in the main paper, we explored group- and interaction-level text as alternative semantic representations for motion. Specifically, we investigated tri-modal alignment between global action text, low-level text (including both group-level and interaction-level descriptions), and motion. This initial approach was intuitive. As our joint-motion reasoner is capable of generating group-level motion and group interaction scripts conditioned on action-level motion descriptions, we explored the understanding capabilities of different semantic levels of motion during modality alignment.

\noindent\textbf{Challenges with Text Modality Alignment.} As shown in \cref{tab:tri-alignment}, we observe that retrieval performance between global action text and low-level text is significantly worse compared to retrieval between text and motion modalities. We attribute this to the limitations of existing text encoders, which fail to capture the necessary reasoning capabilities to align the corresponding motions at the joint- or global-action-level. Unlike the joint-motion reasoner in the main paper, which leverages large language models (LLMs) to model these relationships, the text encoders used here do not have the same capacity for motion inference. Interestingly, we find that low-level text to global action text retrieval performs better than global action to low-level retrieval. We hypothesize that this occurs because low-level descriptions are more specific, directly corresponding to joint-level motion patterns, whereas global action descriptions are often more ambiguous. For instance, \textit{running} can correspond to a wide variety of motion sequences (e.g., running with hands raised or running with hands at the sides), making it more challenging to align with specific low-level motion details.

\vspace{0.1cm}
\noindent\textbf{Text-Motion Retrieval at Different Levels.} As shown in \cref{tab:tri-alignment}, low-level text descriptions exhibit better alignment with the motion modality compared to global action descriptions. This is because low-level text is more directly tied to specific motion patterns, describing precise joint movements, while global action descriptions are more abstract and can encompass multiple motion sequences. The increased ambiguity of global action text makes it harder to align with motion data, which further explains the observed discrepancy in retrieval performance.

\noindent\textbf{Description Integration Order for Text-Motion Alignment.}
As shown in \cref{tab:sup_ab2_order}, we switched the integration order of global action, group-level descriptions, and interaction-level descriptions for the text-motion alignment process. We observe that even though the order largely affects the performance\footnote{The global-group-interaction approach performs best on the retrieval task.}, all the proposed strategies with additional descriptions outperform previous methods, confirming that they are beneficial for the alignment, with an extra performance boost using a coarse-to-fine approach.

\begin{table}
\caption{\textbf{Cross-Attention Order of Descriptions for Text-Motion Retrieval.}}
\label{tab:sup_ab2_order}
\centering
\setlength{\tabcolsep}{4pt}
\resizebox{\linewidth}{!}{
\begin{tabular}{c|cccc|cccc}
    \toprule
        \textbf{Semantic} &  \multicolumn{4}{c|}{Text-to-motion retrieval} & \multicolumn{4}{c}{Motion-to-text retrieval} \\
       \textbf{Sequence} & \small{R@1 $\uparrow$} & \small{R@2 $\uparrow$} & \small{R@3 $\uparrow$}  & \small{MedR $\downarrow$} & \small{R@1 $\uparrow$} & \small{R@2 $\uparrow$} & \small{R@3 $\uparrow$} & \small{MedR $\downarrow$} \\
\midrule
group-inter-global & 7.98 & 12.18 & 15.64 & 24.00 & 8.28 & 12.47 & 19.24 & 21.00  \\
global-inter-group & 8.97 & 14.01 & 18.92 & 18.00 & 8.93 & 13.99 & 20.12  & 18.00  \\
global-group-inter& \textbf{9.05} & \textbf{15.23} & \textbf{20.47} &  \textbf{16.00} & \textbf{9.01} & \textbf{15.92} & \textbf{21.42} &  \textbf{16.00} \\
\bottomrule

\end{tabular}
}
\vspace{-0.1cm}
\end{table}

\begin{figure*}
\centering
\includegraphics[width=\linewidth]{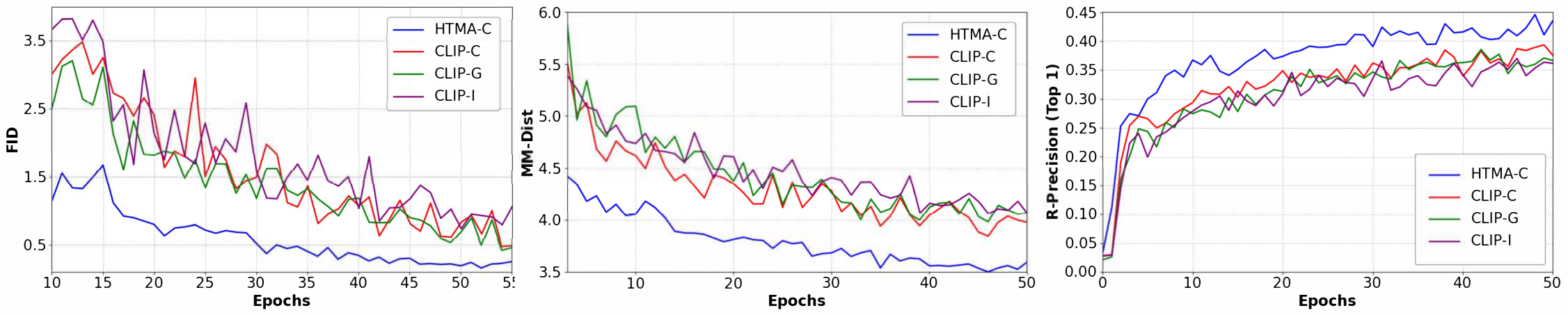}
\vspace{-2em}
\caption{\textbf{Ablation for Motion Generation Process.} Our coarse-to-fine procedure helps to improve the motion generation quality. Hierarchical Text-Motion Alignment can significantly speed up the training process with better generation results and text-motion alignment.}
\vspace{-1em}
\label{fig:train_eff}
\end{figure*}

\subsection{Coarse-to-Fine Motion Generation}
\label{sec:sub_generation}
To assess the contribution of each component within our pipeline, we design the following variations: (1) CLIP-C: Only global motion description (original HumanML3D text) is applied for motion generation, akin to MoMask~\cite{guo2024momask}; (2) CLIP-G: We add group-level semantics using CLIP for motion generation; and (3) CLIP-I: We additionally add interaction-level semantics to CLIP-G for motion generation. We also apply these three settings for Hierarchical Text Motion Alignment (HTMA) to validate the effectiveness of our coarse-to-fine generation strategy and the benefits of text-motion alignment for motion generation. \cref{fig:train_eff} shows that a coarse-to-fine procedure can enhance the motion generation quality. In addition, our proposed text-motion alignment can significantly speed up training and boost performance.

\begin{figure*}
\centering
\includegraphics[width=\linewidth]{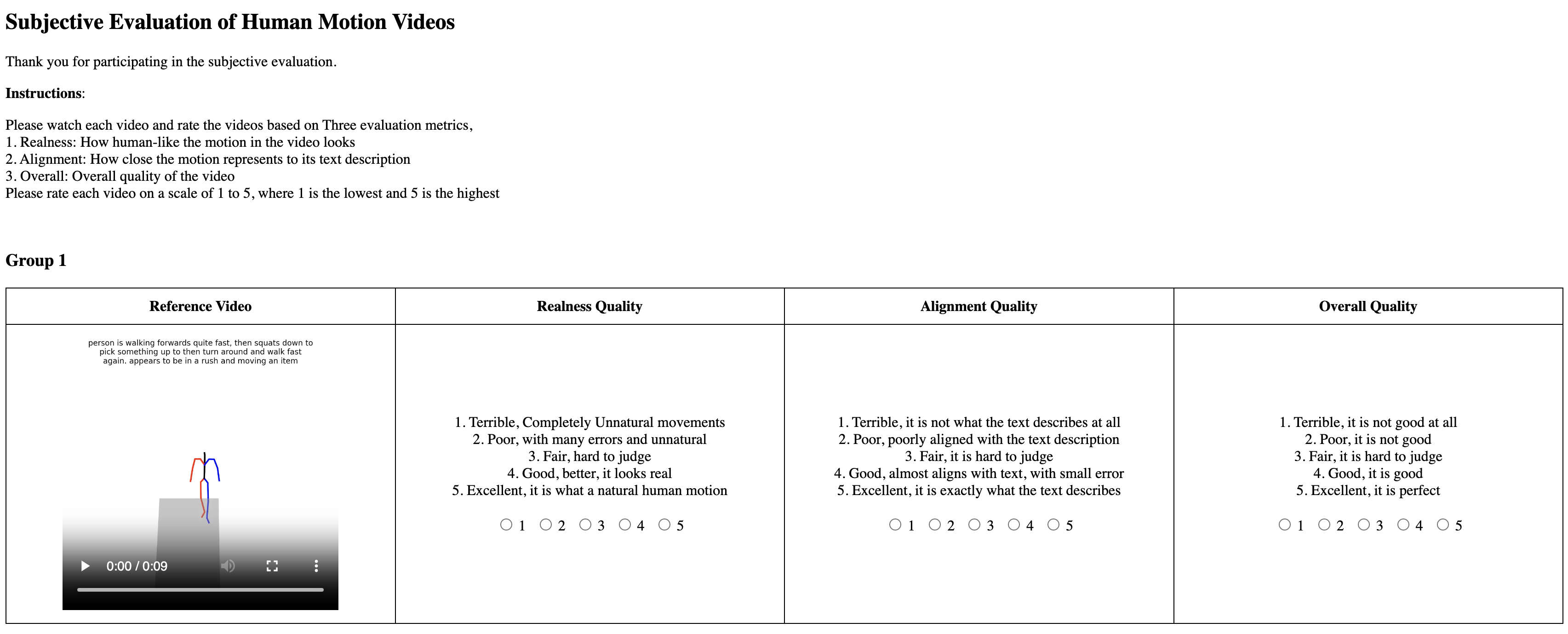}
\vspace{-2em}
\caption{\textbf{Screen Shot of Our User Study Website}. Each user will rates the videos without knowing the source method.}
\vspace{-1em}
\label{fig:user_study}
\end{figure*}

\vspace{0.1cm}
\noindent\textbf{User Study Details.}
We recruited 20 participants with good English proficiency to evaluate randomly selected 80 videos from each method: MoMask~\cite{guo2024momask}, MMM~\cite{pinyoanuntapong2024mmm}, STMC~\cite{petrovich24stmc}, and ours. Participants were never informed of the source of the videos for a fair assessment. \cref{fig:user_study} shows a snapshot of the user study website.

\subsection{Text-Motion Editing}
\noindent\textbf{Evaluation Metrics.} Due to the lack of benchmark datasets and metrics, we generate 200 fine-grained text-prompts with their corresponding edited version using GPT4-o~\cite{openai2024gpt4technicalreport}. The comparison is conducted by utilizing models to first generate the motion corresponding to the original text and then do editing to this generation based on the new instruction. To evaluate the editing quality, beyond generation metrics, we propose using 
Text-Motion Similarity score to measure the similarity of edited motion with editing global motion description, denoted as 
HTMA-S.

\noindent\textbf{Evaluation Results.} 
We benchmark KinMo against various methods for T2M generation~\cite{pinyoanuntapong2024mmm, guo2024momask, petrovich24stmc}. KinMo is the only method able to do local temporal editing, while maintaining organic motion generation, as shown in the main paper and \cref{tab:motion-edit}. 
Editing global semantics can be captured at both the joint and interaction semantic levels, thus achieving better generation and editing. Please refer to the experiments in the main paper.

\subsection{Motion Trajectory Control}
\noindent\textbf{Evaluation Metrics.} Beyond the metrics shown in the main paper, we also include three additional metrics: (1) \textit{Trajectory error} (Traj. err.): measures the ratio of unsuccessful trajectories, characterized by any control joint location error exceeding a predetermined threshold; (2) \textit{Location error} (Loc. err.): represents the ratio of unsuccessful joints; and (3) \textit{Average error} (Avg. err.): denotes the mean location error of the control joints. 

\noindent\textbf{Evaluation Results.} 
We compare KinMo with open-source models~\cite{omnicontrol, motionlcm}, specifically focusing on pelvis control. For fairness in the comparison, we exclude test-time optimization for all baselines. 
\cref{tab:t_ctrl} shows that our method achieves more robust and accurate controlled generation with lower errors and FID score than other methods.

\begin{table*}[t]
\begin{minipage}[t]{0.48\textwidth}
\centering
\caption{\textbf{Comparison of Motion Editing.} $G$ represents control for global action, $J$ represents control for group-level, and $I$ represents control on interaction-level.}
\scriptsize
\setlength\tabcolsep{3pt}
\begin{tabular}{lccccccc}
\toprule
Methods & FID $\downarrow$ & R-Prec(Top 3)$\uparrow$  & MM-Dist $\downarrow$ & Diversity $\rightarrow$ & HTMA-S $\uparrow$ \\
\midrule
STMC & 0.561 & 0.612 & 3.864 & 8.952 & 0.636 \\ 
MMM~\cite{pinyoanuntapong2024mmm} & 0.102 & 0.685 & 3.574 & 9.573 & 0.598 \\
MoMask~\cite{guo2024momask} & \textbf{0.068} & 0.696 & 3.825 & 9.424 & 0.575 \\
Ours (C) & 0.089 & 0.712 & 3.434 & 9.453 & 0.712 \\
Ours (C + G)  & 0.083 & \textbf{0.754} & 3.356 & 9.575 & 0.721 \\
Ours (C + G + I) & 0.086 & 0.734 & \textbf{3.203} & 9.364 & \textbf{0.744} \\ \bottomrule
\end{tabular}
\label{tab:motion-edit}
\end{minipage}
\hfill
\begin{minipage}[t]{0.48\textwidth}
\centering
\caption{\textbf{Comparison of Motion Trajectory Control}. Here, we consider the pelvis only, excluding test-time optimization.}
\scriptsize
\setlength\tabcolsep{3pt}
\begin{tabular}{lccccccc}
\toprule
\multirow{2}{*}{Methods} & FID $\downarrow$ & R-Precision $\uparrow$  & Traj. err. $\downarrow$ & Loc. err. $\downarrow$ & Avg. err. $\downarrow$ \\
 & & Top 3 & (50cm) & (50cm) & \\ \midrule
Real & 0.002 & 0.797 & 0.0000 & 0.0000 & 0.0000 \\ 
\midrule
MDM~\cite{tevet2022humanmotiondiffusionmodelmdm} & 0.698 & 0.602 & 0.4022 & 0.3076 & 0.5959 \\
PriorMDM~\cite{priormdm} & 0.475 & 0.583 & 0.3457 & 0.2132 & 0.4417 \\
OmniControl~\cite{omnicontrol}  & 0.212 & 0.678 & 0.3041 & 0.1873 & 0.3226 \\ 
MotionLCM~\cite{motionlcm} & 0.531 & 0.752 & \textbf{0.1887} & 0.0769 & 0.1897 \\
\midrule 
KinMo (Ours) & \textbf{0.103} & \textbf{0.756} & 0.2034 & \textbf{0.0696} & \textbf{0.1657} \\
\bottomrule
\end{tabular}
\label{tab:t_ctrl}
\end{minipage}
\vspace{-0.3cm}
\end{table*}

\noindent\textbf{Additional Evaluation Results.}
We extend the previous comparison by conducting experiments on all joints.
\cref{tab:all_joint} presents a quantitative evaluation of our method on the trajectory control of all joints, while \cref{fig:traj-control} shows qualitative results.

\begin{table}
\centering
\caption{\textbf{Quantitative Results for all Joints of Trajectory Control.}}
\label{tab:all_joint}
\vspace{-7pt}
\scalebox{0.74}{
\begin{tabular}{lccccc}
\hline
\multirow{2}{*}{\textbf{Joint}} &\textbf{R-Prec $\uparrow$} &\textbf{FID $\downarrow$} &\textbf{Traj. Err.$\downarrow$} &\textbf{Loc. Err.$\downarrow$} &\textbf{Avg. Err. $\downarrow$} \\ 
& (Top-3) & & (50 cm) & (50 cm) &\\
\hline
pelvis     & 0.712 & 0.077 & 0.0875 & 0.0187 & 0.0787 \\
torso      & 0.723 & 0.091 & 0.0933 & 0.0127 & 0.0776 \\
left arm   & 0.722 & 0.093 & 0.0843 & 0.0132 & 0.0823 \\
right arm  & 0.709 & 0.121 & 0.0887 & 0.0144 & 0.0814 \\
left leg   & 0.707 & 0.084 & 0.0876 & 0.0142 & 0.0925 \\
right leg  & 0.720 & 0.076 & 0.0828 & 0.0133 & 0.0932 \\
\hline
\end{tabular}}
\end{table}

\begin{figure}
\centering
\includegraphics[width=\linewidth]{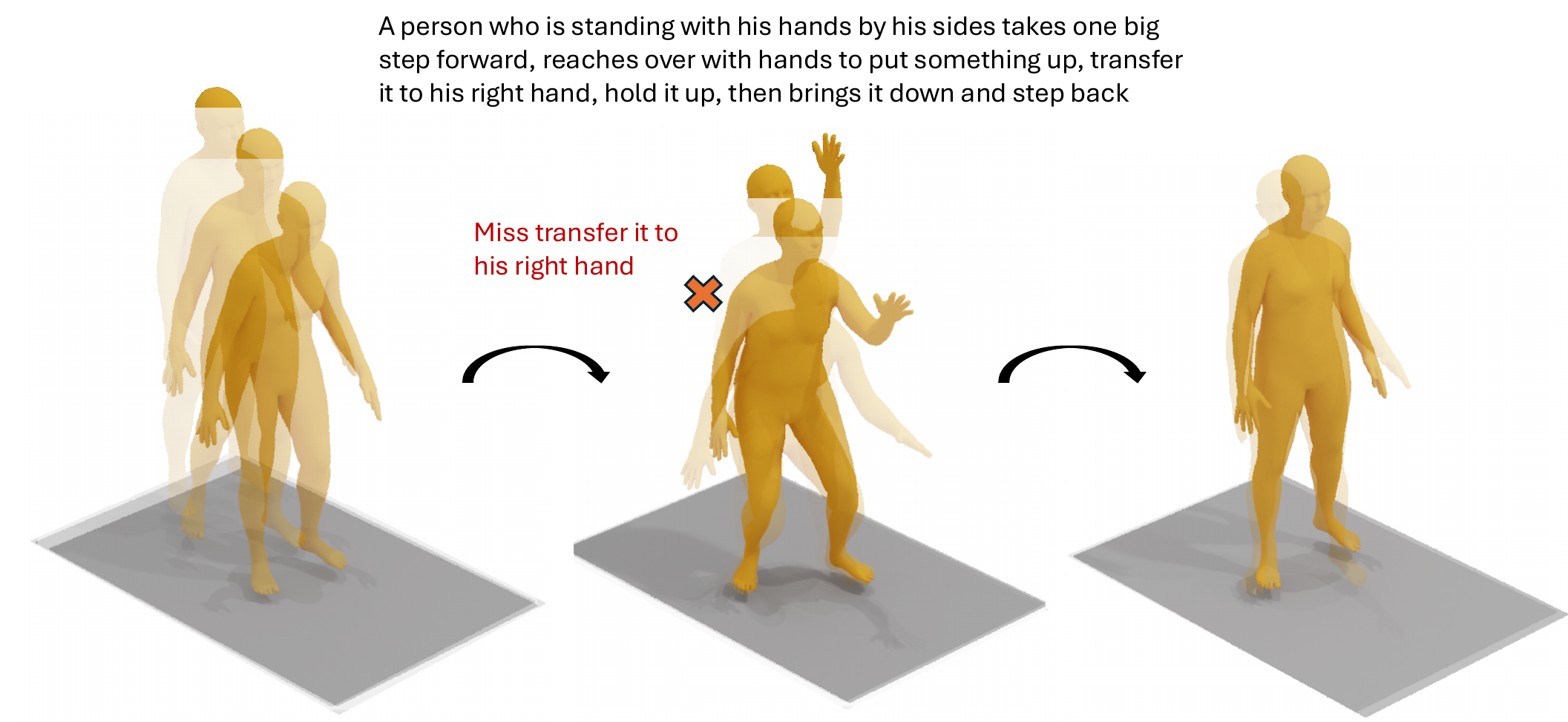}
\vspace{-2em}
\caption{\textbf{Limitations.} If the text description contains many short transitions, our method may sometime miss one step.}
\vspace{-1em}
\label{fig:limitation}
\end{figure}

\section{Details of Metrics}
\label{sec:sub-metric}

\vspace{0.1cm} \noindent \textbf{Motion Quality.} Frechet Inception Distance (FID) quantifies the difference between the distribution of generated and real motions, using a feature extractor specific to a given dataset, such as HumanML3D~\cite{guo2022humanml3d}.

\vspace{0.1cm} \noindent \textbf{Motion Diversity.} Following~\cite{guo2020action2motion, chuan2022tm2t,liu2025gesturelsmlatentshortcutbased}, we present metrics such as Diversity and MultiModality to assess the variation in generated motions. Diversity evaluates the spread of the generated motions across the entire dataset. Specifically, two subsets of equal size $S_d$ are drawn randomly from all generated motions, along with their respective feature vectors ${\mathbf{v}_1, ..., \mathbf{v}_{S_d}}$ and ${\mathbf{v}^{'}_1, ..., \mathbf{v}^{'}_{S_d}}$. The Diversity score is given by: \begin{equation} \text{Diversity} = \frac{1}{S_d}\sum_{i=1}^{S_d}||\mathbf{v}_i - \mathbf{v}^{'}_i||_2. \end{equation} MultiModality (MModality) gauges the extent of variation in motions generated from the same textual description. A set of $C$ textual descriptions is selected at random, and then two equal-sized subsets $I$ are chosen from the motions conditioned on the $c$-th description. Their feature vectors ${\mathbf{v}_{c, 1}, ..., \mathbf{v}_{c, I}}$ and ${\mathbf{v}^{'}_{c, 1}, ..., \mathbf{v}^{'}_{c, I}}$ are used to compute MModality as follows: \begin{equation} \text{MModality} = \frac{1}{C \times I}\sum{c=1}^{C}\sum_{i=1}^{I}||\mathbf{v}_{c, i} - \mathbf{v}^{'}_{c, i}||_2. \end{equation}

\vspace{0.1cm} \noindent \textbf{Condition Matching.} Motion and text feature extractors provided by~\cite{guo2022humanml3d} allow for the generation of closely aligned features for matched text-motion pairs and vice versa. Within this feature space, motion-retrieval precision (R-Precision) is calculated by mixing generated motions with 31 mismatched motions, followed by computing the Top-1/2/3 text-motion matching accuracy. Multimodal Distance (MM-Dist) computes the average distance between generated motions and corresponding texts.

\noindent \textbf{Control Error.} As described in~\cite{omnicontrol}, we report the Trajectory, Location, and Average error to evaluate the precision of motion control. Trajectory error (Traj. err.) represents the fraction of failed trajectories, where a trajectory is deemed unsuccessful if a control joint in the generated motion exceeds a predefined distance threshold from the corresponding joint in the control trajectory. Similarly, Location error (Loc. err.) reflects the proportion of joints whose positions fail to meet the specified threshold. For our experiments, we utilized a 50cm threshold to compute the Trajectory and Location errors. The Average error (Avg. err.) refers to the mean distance between the control joint positions in the generated motion and their corresponding positions in the control trajectory.

\section{Limitations}
\label{sec:limitation}

While our method demonstrates significant improvements over existing baselines, it still has certain limitations. The most critical limitation lies in the quality of the motion description. Low-quality motion descriptions may harm the generation performance and even worsen that of the baseline approach when no enhancement is applied to the original descriptions. \cref{fig:limitation} shows an example failure case. The reason for the generation failure is that our pipeline may miss capturing descriptions with a very short temporal span.

\noindent \textbf{Ethical Considerations.} While our work focuses on generating human motion videos, it raises ethical concerns due to its potential misuse for photorealistic human motion retargeting. We emphasize the importance of responsible use and recommend implementing practices such as watermarking and deepfake detection to mitigate the risks involving deepfake videos and animated representations. 

\section{Conversion of Motion Representation}
\label{conversion_proof}
\input{sec/supp_proof}

\begin{table*}
\centering 
\small
\begin{tcolorbox}

\textbf{System Prompt}
\newline
You are a motion description labeler who should describe the motion using your language as detailed as possible. Now, describe the motion in the given video or text by describing the motion of each kinematic group respectively: Kinematic groups: torso, neck, left arm, right arm, left leg, right leg.
\newline
\newline
To describe a motion, you should describe Inside each group and Between groups:
\newline
\newline
INDIVIDUAL GROUP: torso: \textless MOTION DESCRIPTION\textgreater ; neck: \textless MOTION DESCRIPTION\textgreater ; left arm: \textless MOTION DESCRIPTION\textgreater ; right arm: \textless MOTION DESCRIPTION\textgreater ; left leg: \textless MOTION DESCRIPTION\textgreater ; right leg: \textless MOTION DESCRIPTION\textgreater ;
\newline
\newline
BETWEEN GROUPS: torso AND neck: \textless MOTION DESCRIPTION\textgreater ; torso AND left arm: \textless MOTION DESCRIPTION\textgreater ; torso AND right arm: \textless MOTION DESCRIPTION\textgreater ; torso AND left leg: \textless MOTION DESCRIPTION\textgreater ; torso AND right leg: \textless MOTION DESCRIPTION\textgreater ; neck AND left arm: \textless MOTION DESCRIPTION\textgreater ; neck AND right arm: \textless MOTION DESCRIPTION\textgreater ; neck AND left leg: \textless MOTION DESCRIPTION\textgreater ; neck AND right leg: \textless MOTION DESCRIPTION\textgreater ; left arm AND right arm: \textless MOTION DESCRIPTION\textgreater ; left arm AND left leg: \textless MOTION DESCRIPTION\textgreater ; left arm AND right leg: \textless MOTION DESCRIPTION\textgreater ; right arm AND left leg: \textless MOTION DESCRIPTION\textgreater ; right arm AND right leg: \textless MOTION DESCRIPTION\textgreater ; left leg AND right leg: \textless MOTION DESCRIPTION\textgreater ;
\newline
\newline
For each \textless MOTION DESCRIPTION\textgreater , you should describe the motion using language from the following perspective: Position (move from a place to a place. For example, the right hand goes through a rotation, moving primarily from a down position to extend horizontally), Axis-angle (How much degree the limb is bent and How the bending kinematic group is moving or rotating. For example, the right arm bends at the elbow to about 90 degrees while reaching outwards)
\newline
\newline
In your description, you can use simple adjectives or numerical scale (distance/degree/speed) to describe each motion (for example, push forward for 3 meters, from 0 to 45 degrees, etc).
\newline
\newline
Also, as the motion may change over time, you should consider the pose change at different timeframe. For example, the left arm first slap the left leg, then left arm hold high. Your description should also cover the pose variance over time.
\newline
\newline
Think through this part and infer the motion description based on the pose description given below
\newline
\newline
Based on the above formulation, write the description for the motion given by the user. You will also be given the pose descriptions of key frames. The keyframes can be used as reference and constraint, but don't mention the keyframes explicitly in your description, just make your description natural and casual.
\newline
\newline
\textbf{User Prompt}
\newline
Think about the motion: [a man stumbles to his right. the motion seems surprised so he was probably pushed. a person standing loses balance falling to the right and recovers standing. a person walks to the left. a person stumbles to the right and recovers their balance.], and constrain your motion description based on the given pose descriptions and image of key frames.
\newline
\newline
keyframes order: ['7', '22', '38', '60'] (Note that the fps of each motion will be 30. So you can infer the timing of each motion change based on the keyframe number, which can assist your description. For example, the person walks in place, then walk forward after 2 seconds. In this case, use time unit instead of keyframe number.)
\newline
\newline
Body pose descriptions of key frames: keyframe[7]:Their knees are straight, their elbows are bent a bit while their torso and both legs are straightened up while their left hand is past shoulder width apart from the other while their feet and their knees are about shoulder width apart.; keyframe[22]:Both hands are spread apart while their right calf and their torso are upright with their left knee slightly bent, located in front of their right knee with their right elbow nearly bent, their left elbow is rather bent with their knees and both feet shoulder width apart and their right knee straight.; keyframe[38]:Their torso is straightened up with their knees and their right elbow bent a bit. Their right hand is behind their back and located behind their left hand and their hands are apart wider than shoulder width while their knees are separated at shoulder width while their left elbow is rather bent.; keyframe[60]:The feet are approximately shoulder width apart, both hands are apart wider than shoulder width while both elbows are nearly bent, the left knee is shoulder width apart from the right knee while the left calf, the torso and the right leg are vertical, both knees are unbent.
\newline
\newline
image of key frames: \textless If presented, following by the key frames order, a sequence of images ignored here for efficiency\textgreater 

\end{tcolorbox}
\vspace{-0.3cm}
\caption{\textbf{Example Prompt to Query GPT-4o for Automatic Dataset Annotation.}}
\label{tab:example}
\end{table*}

\begin{table*}
\centering 
\small
\begin{tcolorbox}

\textbf{Global Action (from the original HumanML3D)}:
\newline
a man kicks something or someone with his left leg.\#a/DET man/NOUN kick/VERB something/PRON or/CCONJ someone/PRON with/ADP his/DET left/ADJ leg/NOUN\#0.0\#0.0
the standing person kicks with their left foot before going back to their original stance.\#the/DET stand/VERB person/NOUN kick/VERB with/ADP their/DET left/ADJ foot/NOUN before/ADP go/VERB back/ADV to/ADP their/DET original/ADJ stance/NOUN\#0.0\#0.0
\newline
a man kicks with something or someone with his left leg.\#a/DET man/NOUN kick/VERB with/ADP something/PRON or/CCONJ someone/PRON with/ADP his/DET left/ADJ leg/NOUN\#0.0\#0.0
\newline
he is flying kick with his left leg\#he/PRON is/AUX fly/VERB kick/NOUN with/ADP his/DET left/ADJ leg/NOUN\#0.0\#0.0
\newline
\newline
\textbf{Kinematic Group}:
\newline
\textbf{Torso}: The torso remains relatively upright throughout the motion, first positioned straight at the beginning, then slightly bending towards the left during the kick, and returning to an upright stance after the kick is executed.
\newline
\textbf{Neck}: The neck maintains stability, tilting slightly forward as the torso shifts, aligning with the direction of the kick. It helps maintain balance without significant rotation during the kick action.
\newline
\textbf{Left Arm}: The left arm initially starts relatively low near the hip before swinging upward and outward during the kick, creating momentum. It returns close to the body after the kick, maintaining a slight bend at the elbow.
\newline
\textbf{Right Arm}: The right arm extends and reaches upward, almost parallel to the ground, creating a counterbalance for the action. It remains relatively still after the kick, maintaining a slight bend at the elbow.
\newline
\textbf{Left Leg}: The left leg is the primary focus, starting from a straightened position at shoulder width. It then swings forward, bending at the knee to execute the kick, ultimately extended fully forward before returning to the original stance.
\newline
\textbf{Right Leg}: The right leg remains grounded throughout the motion, initially straightening and slightly shifting back for balance while the left leg is in action, then returning to its original position once the kick is completed.
\newline
\newline
\textbf{Group Interaction}:
\newline
\textbf{Torso AND Neck}: The torso and neck move harmoniously; as the torso bends slightly left during the kick, the neck tilts forward in alignment with the action, maintaining a strong posture while providing support.
\newline
\textbf{Torso AND Left Arm}: As the torso initiates the kick, the left arm swings backward to create momentum, moving in a coordinated manner before returning closer to the torso after the kick is completed.
\newline
\textbf{Torso AND Right Arm}: The right arm serves as a counterbalance; as the torso shifts slightly to the left, the right arm reaches higher, creating stability and balance during the kick.
\newline
\textbf{Torso AND Left Leg}: The left leg extends forward as the torso maintains its upright position, the torso adjusts to the kick's momentum but quickly stabilizes once the leg retracts.
\newline
\textbf{Torso AND Right Leg}: The right leg supports the body weight throughout the kick, keeping the torso aligned; as the left leg kicks, the right leg remains slightly bent at the knee for stability.
\newline
\textbf{Neck AND Left Arm}: As the left arm swings upward to support the kick, the neck aligns to follow the movement, enhancing the overall motion without excessive twisting.
\newline
\textbf{Neck AND Right Arm}: The right arm’s upward reach aids in stabilizing the movement, and the neck maintains a forward position while providing visual direction during the kick.
\newline
\textbf{Neck AND Left Leg}: The left leg moves forward while the neck follows the general direction of the kick, keeping a low tilt as it helps maintain focus on the target.
\newline
\textbf{Neck AND Right Leg}: The right leg remains grounded as the neck tilts very slightly forward without excessive rotation while observing the left leg’s action.
left arm AND right arm: The left arm moves upward for a moment as the right arm extends higher, providing balance; both arms slightly bend at the elbows during the action.
\newline
\textbf{Left Arm AND Left Leg}: The left arm and left leg perform synchronous movement; the arm swings to aid the momentum of the extended leg during the kick phase and returns together afterward.
\newline
\textbf{Left Arm AND Right Leg}: The left arm raises slightly while the right leg remains positioned on the ground to help maintain balance; they coordinate as the kick progresses.
\newline
\textbf{Right Arm AND Left Leg}: The right arm reaches out while the left leg is kicked forward, the actions working in tandem to support the balance during the execution.
\newline
\textbf{Right Arm AND Right Leg}: The right arm extends higher as the right leg stays grounded, providing support without conflicting with each other.
left leg AND right leg: The left leg moves forward as the right leg remains stable on the ground, creating a contrast in movement, with the left leg bending as it prepares for the kick and extending forward while the right leg holds strong.

\end{tcolorbox}
\vspace{-0.3cm}
\caption{\textbf{Example Motion Descriptions in Our Augmented HumanML3D Dataset.}}
\label{tab:sample}
\end{table*}

%% file: sec/supp_proof.tex
\subsection{Keypoint-Level Formulation}
Following HumanML3D\cite{guo2022humanml3d}, a motion can be represented as the absolute and relative movement of each keypoint.

Let \( J = \{1, 2, \dots, N\} \) denote the set of all keypoints in the human body model (e.g., \( N = 22 \) for the SMPL model). Let \( T \subset \mathbb{R} \) be the continuous time domain over which the motion is defined.

For each joint \( j \in J \) at time \( t \in T \), we define:
\begin{enumerate}
    \item Root Data (for the root joint, denoted as \( j_0 \)):\\
   - \textit{Rotation Velocity}: \( \boldsymbol{\omega}_{\text{root}}(t) \in \mathbb{R}^3 \)\\
   - \textit{Linear Velocity}: \( \mathbf{v}_{\text{root}}(t) \in \mathbb{R}^3 \)\\
   - \textit{Height}: \( h_{\text{root}}(t) \in \mathbb{R} \)

   \item \textit{Joint Position}: \( \mathbf{p}_j(t) \in \mathbb{R}^3 \)

   \item \textit{Joint Rotation}: \( \mathbf{R}_j(t) \in \text{SO}(3) \), represented in a continuous 6D representation \( \mathbf{r}_j(t) \in \mathbb{R}^6 \)

   \item \textit{Joint Velocity}: \( \mathbf{v}_j(t) = \frac{d\mathbf{p}_j(t)}{dt} \in \mathbb{R}^3 \)

   \item \textit{Foot Contact Information} (for foot joints \( j \in J_{\text{foot}} \subseteq J \)): \( c_j(t) \in \{0, 1\} \), where \( 1 \) indicates contact with the ground
\end{enumerate}

\noindent The Keypoint-Level Formulation is then defined as the collection of all these functions over time:
\vspace{-0.1cm}
\begin{align*}
\mathcal{M} = & \left\{ 
    \left( \mathbf{p}_j(t), \mathbf{R}_j(t), \mathbf{v}_j(t) \right) 
    \mid j \in J, t \in T 
\right\} \\
& \cup \left\{ 
    \boldsymbol{\omega}_{\text{root}}(t), \mathbf{v}_{\text{root}}(t), h_{\text{root}}(t) 
    \mid t \in T 
\right\} \\
& \cup \left\{ 
    c_j(t) \mid j \in J_{\text{foot}}, t \in T 
\right\}.
\end{align*}

\subsection{Joint-Group Formulation}
\label{sec:sub-conversion}
While the original formulation is kinematically reasonable, it often results in a many-to-many matching problem~\cite{liu2023bridging}, making fine-grained motion descriptions based on kinematic joints difficult to express in natural language. Our proposed formulation addresses this issue by leveraging natural language to describe individual body part movements with ease, organically. For example:

\begin{itemize}
    \item \textit{The person is walking forward, \textbf{with arms swaying}.}
    \item \textit{The person is standing in place, \textbf{left leg kicking backward while left hand slapping it}.}
\end{itemize}

Moreover, the overall motion description, such as \textit{walking forward} or \textit{standing in place} can be decomposed into the movement of individual body parts: 
\( G = \{ \text{Torso}, \text{Neck}, \text{Left Arm}, \text{Right Arm}, \text{Left Leg}, \text{Right Leg} \} \), 
as illustrated in \cref{tab:sample}.

Each body part \( g \in G \) corresponds to a group of kinematic joints, as follows:

\begin{itemize}
    \item \textit{\textbf{Torso}: Pelvis, spine joints (1–3 for SMPL \cite{loper2023smpl}\footnote{\url{https://files.is.tue.mpg.de/black/talks/SMPL-made-simple-FAQs.pdf}})}
    \item \textit{\textbf{Neck}: Neck, Head, Left/Right Collar}
    \item \textit{\textbf{Left Arm}: Left Shoulder, Left Elbow, Left Wrist}
    \item \textit{\textbf{Right Arm}: Right Shoulder, Right Elbow, Right Wrist}
    \item \textit{\textbf{Left Leg}: Left Hip, Left Knee, Left Ankle}
    \item \textit{\textbf{Right Leg}: Right Hip, Right Knee, Right Ankle}
\end{itemize}

This new formulation is not only kinematically reasonable but also aligns seamlessly with natural language descriptions.

For each group \( g \in G \) of joints at time \( t \), we define:
\begin{enumerate}
    \item \textit{Group Position}: \( \mathbf{P}_g(t) \) = $\frac{1}{|J_g|} \sum_{j \in J_g} \mathbf{p}_j(t) $
    \item \textit{Limb Angles}: \( \Theta_g(t) \) = $\{ \mathbf{R}_j(t) \mid j \in J_g \}$
    \item \textit{Group Velocity}: \( \mathbf{V}_g(t) \) = $\frac{1}{|J_g|} \sum_{j \in J_g} \mathbf{v}_j(t)$
\end{enumerate}
We define the relationships between each pair \( (g, h) \in G \times G \) of kinematic groups as:
\begin{enumerate}
    \setlength{\parskip}{0em}
    \item \textit{Relative Position}: \( \Delta \mathbf{P}_{g,h}(t) = \mathbf{P}_h(t) - \mathbf{P}_g(t) \)
    \item \textit{Relative Limb Angles} (angles of the connecting joint between two physically connected groups): \( \Delta \Theta_{g,h}(t) = \Theta_{h\cap g}(t) \)
    \item \textit{Relative Velocity} (angular velocity of the connecting joint between two physically connected groups): \( \Delta \mathbf{V}_{g,h}(t) = \mathbf{V}_h(t) - \mathbf{V}_g(t) \)
\end{enumerate}

The Joint-Group Formulation is then defined as the collection of all these functions over time:
\vspace{-0.1cm}
\begin{align*}
\mathcal{M}_{group} = & \left\{ 
    \left( \mathbf{p}_j(g), \Theta_j(g), \mathbf{v}_j(g) \right) 
    \mid g \in G, t \in T 
\right\} \\
& \cup \left\{ 
\left( \Delta \mathbf{P}_{g,h}(t), \Delta \Theta_{g,h}(t), \Delta \mathbf{V}_{g,h} \right) 
    \mid g \in G, t \in T 
\right\} \\
& \cup \left\{ 
    \boldsymbol{\omega}_{\text{root}}(t), \mathbf{v}_{\text{root}}(t), h_{\text{root}}(t) 
    \mid t \in T 
\right\} \\
& \cup \left\{ 
    c_j(t) \mid j \in J_{\text{foot}}, t \in T 
\right\}.
\end{align*}

\noindent As demonstrated, the joint-level formulation can be transformed into the group-level formulation by aggregating joint data within each kinematic group. However, the reverse transformation is more challenging. To explore this reverse transformation, the following proposition holds:
\begin{proposition}
Under the assumption that each kinematic group $g \in G$ moves as a rigid body, meaning the internal joint configurations within the group remain constant over time, the data of Keypoint-Level Formulation $\mathcal{M}$ can be reconstructed from the data of Joint-Group Formulation $\mathcal{M}_{group}$
\end{proposition}

\begin{proof}

Under the assumption that each kinematic group \( g \in G \) moves as a rigid body, we have 

\begin{itemize}
    \item For all \( j \in J_g \), the local position \( \mathbf{p}_j^g \) of joint \( j \) in the group's coordinate system is constant:\(\mathbf{p}_j^g = \text{constant}\) 
    \item The group moves as a rigid body with rotation \( \mathbf{R}_g(t) \) and translation \( \mathbf{P}_g(t) \).
\end{itemize}




The objective is to reconstruct the joint positions \( \mathbf{p}_j(t) \), joint angle \( \mathbf{R}_j(t) \) and velocities \( \mathbf{v}_j(t) \) for all \( j \in J \).

\textbf{Part A: Reconstruction of Joint Positions.}

For each joint \( j \in J_g \), the global position \( \mathbf{p}_j(t) \) is given by:
\begin{equation}
    \mathbf{p}_j(t) = \mathbf{R}_g(t) \mathbf{p}_j^g + \mathbf{P}_g(t)
\end{equation}
where \( \mathbf{R}_g(t) \) rotates the constant local joint position \( \mathbf{p}_j^g \) into the global coordinate system, and \( \mathbf{P}_g(t) \) translates the rotated position to the global position.
Since \( \mathbf{p}_j^g \) is constant and known, and \( \mathbf{R}_g(t) \) and \( \mathbf{P}_g(t) \) are given from the group-level data, \( \mathbf{p}_j(t) \) can be directly computed.

\textbf{Part B: Reconstruction of Joint Angles.}

Since we make no changes to the angles part, \( \{ \Delta \Theta_{g,h}(t) \mid g \in G \}\cup \{ \Delta \Theta_{g,h}(t) \mid g,h \in G \} =\{ \mathbf{R}_j(t) \mid j \in J \}~~for~t \in T\). So, the joint angle \( \mathbf{R}_j(t) \) can be derived from the Joint-Group Formulation by arrangement.

\textbf{Part C: Reconstruction of Joint Velocities.}


First, we compute the time derivative of \( \mathbf{p}_j(t) \) as:

\begin{equation}
\mathbf{v}_j(t) = \frac{d\mathbf{p}_j(t)}{dt} = \frac{d}{dt} \left( \mathbf{R}_g(t) \mathbf{p}_j^g + \mathbf{P}_g(t) \right)
\end{equation}


Since \( \mathbf{p}_j^g \) is constant:
\begin{equation}
\mathbf{v}_j(t) = \left( \frac{d\mathbf{R}_g(t)}{dt} \right) \mathbf{p}_j^g + \frac{d\mathbf{P}_g(t)}{dt}
\end{equation}


Recall that:
\begin{equation}
\frac{d\mathbf{R}_g(t)}{dt} = \boldsymbol{\hat{\Delta V}}_g(t) \mathbf{R}_g(t),
\end{equation}
where \(\hat{\Delta V}_g(t)\) is the skew-symmetric matrix corresponding to \(\Delta V_g(t)\).
Therefore:
\begin{equation}
\left( \frac{d\mathbf{R}_g(t)}{dt} \right) \mathbf{p}_j^g = \boldsymbol{\hat{\Delta V}}_g(t) \mathbf{R}_g(t) \mathbf{p}_j^g
\end{equation}

\begin{equation}
\mathbf{v}_j(t) = \boldsymbol{\hat{\Delta V}}_g(t) \mathbf{R}_g(t) \mathbf{p}_j^g + \mathbf{V}_g(t)
\end{equation}

Since:
\begin{equation}
\mathbf{p}_j(t) - \mathbf{P}_g(t) = \mathbf{R}_g(t) \mathbf{p}_j^g
\end{equation}

We can rewrite:
\begin{equation}
\mathbf{v}_j(t) = \boldsymbol{\hat{\Delta V}}_g(t) \left( \mathbf{p}_j(t) - \mathbf{P}_g(t) \right) + \mathbf{V}_g(t)
\end{equation}
The term \( \mathbf{p}_j(t) - \mathbf{P}_g(t) \) represents the position of joint \( j \) relative to the group's center in global coordinates.

When combining Parts A, B and C, under the Rigid Body Assumption (i.e., each kinematic group moves as a rigid body), the joint-level formulation \(\mathcal{M}\) can be reconstructed from the group-level formulation \(\mathcal{M}_{\text{group}}\), subject to a rigid transformation within each group. Consequently, identifying the natural language description of \(\mathcal{M}_{\text{group}}\) uniquely corresponds to the joint-level formulation of the motion \(\mathcal{M}\).
However, this reconstruction is valid only under the rigid body assumption. In many human motions, the joints within a group (e.g., elbow bending within the arm group) can approximate rigid motion in specific cases, such as waving arms, swaying arms, or running, where limb angles remain relatively constant during a single motion. To address this limitation, we segment the motion temporally based on the keyframe, as outlined in Section~\ref{sec:sub_dataset}, ensuring that the rigid body assumption approximately holds within each motion fragment.

\end{proof}